\documentclass{article}

\usepackage{PRIMEarxiv}

\usepackage[utf8]{inputenc} 
\usepackage[T1]{fontenc}    
\usepackage{hyperref}       
\usepackage{url}            
\usepackage{booktabs}       
\usepackage{amsfonts}       
\usepackage{nicefrac}       
\usepackage{microtype}      
\usepackage{lipsum}
\usepackage{fancyhdr}       
\usepackage{graphicx}       
\graphicspath{{media/}}     
\usepackage{cite}
\usepackage{amsmath,amssymb,amsfonts}
\usepackage{algorithmic}
\usepackage{graphicx}
\usepackage{textcomp}
\usepackage[utf8]{inputenc}
\usepackage{url}
\usepackage{color}
\usepackage{placeins}
\usepackage{booktabs}
\usepackage{tabularx}
\usepackage{array}
\usepackage{makecell}
\usepackage{adjustbox}

\title{StutterStigma-TR: A Novel Turkish Dataset and Target-Aware BERT Framework for Stuttering-Related Stigma Detection
}

\author{
  Muhsin D\"{o}lek\\
  Department of Artificial Intelligence Engineering \\
  Department of Speech and Language Therapy\\
  Bahcesehir University \\
  Istanbul, T\"{u}rkiye\\
  \texttt{muhsindolek@gmail.com} \\
   \And
  Arezoo Sadeghzadeh\\
  Department of Artificial Intelligence Engineering \\
  Bahcesehir University \\
  Istanbul, T\"{u}rkiye\\
  \texttt{arezoo.sadeghzadeh@bau.edu.tr} \\
}

\begin{document}
\maketitle

\begin{abstract}
Computational research on disability-related ableist harm remains limited,
and condition-specific resources for stuttering-related stigma are particularly
scarce. This study introduces StutterStigma-TR, a Turkish dataset of 6,019
public X/Twitter posts annotated as Stigmatizing,
Non-stigmatizing/contextual, or Supportive using a speech and language
therapy-informed protocol with independent lived-experience evaluation.
We also propose StutterStigma-BERT, a BERTurk-based target-aware framework
that combines explicit target marking, target-span representation,
target-conditioned context, character-level and handcrafted orthographic
information, and gated discourse cues. The framework is evaluated in
complementary three-class and binary settings using controlled three-seed
ablation analyses and final 20-seed robustness evaluations. Under the common
training protocol, Gated Discourse-OrthoTarget trained with cross-entropy is
selected as the final configuration for both tasks based on internal-validation
Macro-F1. Across 20 random seeds, the final three-class model achieves a mean
Macro-F1 of 73.54\%, while the final binary model achieves 85.17\%
using validation-selected decision thresholds.
Task-adapted harmful-language comparators are additionally evaluated under
the common training protocol, and the proposed model obtains the highest mean Macro-F1 among them. For the primary three-class task, matched
backbone analyses show higher internal-validation Macro-F1 with BERTurk
than with TurkishBERTweet or Turkish Mamba, while substantial
train--validation separation persists across all three backbones. Separate
direct-transfer and zero-shot experiments examine performance without
StutterStigma-TR supervision. Across the task-adapted,
direct-transfer, and zero-shot comparisons, the condition-specific supervised
systems retain the strongest aggregate performance. Overall, the findings
support treating stuttering-related stigma as a context-sensitive
classification problem in which the marked target expression must be interpreted
together with its surrounding discourse.
\end{abstract}

\keywords{Ableism \and BERTurk \and Deep learning \and Datural language processing \and Social media \and Stigma \and Stuttering \and Target-aware modeling}

\section{Introduction}
\label{sec:introduction}
Stuttering involves observable disruptions in speech fluency, including sound or syllable repetitions, prolongations, and blocks, but its impact extends far beyond these speech features. It is also a multidimensional communication experience shaped by anticipation, listener reactions, avoidance, concealment, participation, and social evaluation  \cite{asha_fluency_disorders, tichenor2019stuttering, tichenor_etal_2022}. Consequently, the social meanings attached to stuttering are central to understanding how people who stutter are perceived, evaluated, and treated in everyday interactions. Public attitudes toward stuttering often involve inaccurate and stigmatizing assumptions. People who stutter may be perceived as nervous, insecure, less confident, less competent, or less suitable for communication-demanding roles   \cite{boyle2013assessment, gabel2004measuring, gerlach2018stuttering, ozdemir2011public, louis2011public}. These biased perceptions are deeply rooted in broader processes of stigma and ableism. Stigma involves labelling human differences, linking them to negative stereotypes, separating people through an “us versus them” distinction, and ultimately subjecting individuals to status loss and discrimination  \cite{goffman2009stigma,link2001conceptualizing}. Within an ableist framework, these assumptions reflect a broader communicative norm in which fluent speech is privileged and stuttered speech may be treated as deficient, undesirable, or incompatible with competence  \cite{campbell2009contours, goodley2014dis, hehir2002eliminating}. 

Social media makes stuttering-related attitudes visible in naturally occurring public discourse \cite{fuse2016impact, nicolai2018retrospective, bidakhavidi2025social, ccavdar2026stuttering}.  Posts about stuttering may include direct mockery, written imitation, negative stereotypes, misinformation, exclusion, or assumptions about competence. At the same time, online discourse about stuttering is not uniformly harmful. It may also include self-disclosure, clinical information, awareness messages, advocacy, community support, and criticism of stigma \cite{donaher2014portrayal, ccavdar2025reliable, azios2022twitter, rasoli2025stuttering}. Consequently, a stuttering-related expression must be analyzed in context rather than reflexively interpreted as evidence of stigma. Its discourse function depends on whether it is used to ridicule stuttering, describe it neutrally, report or quote stigmatizing language, or oppose that language. 

This creates a methodological challenge for automated text classification. General-purpose systems for toxicity, hate speech, cyberbullying, or offensive-language detection are not designed to capture all condition-specific forms of disability-related stigma. Such systems may over-rely on obvious linguistic cues such as profanity, hostility, threat, or explicit abuse, while overlooking more subtle expressions of ableist meaning. Conversely, they may over-flag neutral, self-referential, educational, or supportive disability-related language because disability terms become associated with negativity or toxicity \cite{phutane2025cold, venkit_etal_2023, hutchinson2020social}. For stuttering-related discourse, both forms of error are particularly important: subtle stigma may go undetected when it is expressed without overt abuse, whereas supportive or contextual content may be misclassified when it contains terms or language patterns associated with stigma or discrimination.

Computational research specifically addressing disability-related ableist harm remains limited compared with the extensive literature on general toxicity, hate speech, and offensive-language detection. Existing disability-related NLP research has primarily investigated whether sentiment analysis, toxicity detection, and language models reproduce or amplify undesirable associations with disability-related expressions, rather than developing condition-specific approaches for detecting ableist harm. 
More direct computational work on ableist language has only recently begun to emerge. Existing studies suggest that disability-related harm is often context-dependent, shaped by social assumptions and by the relationship between an identity-related expression and the stance adopted toward it. At the condition-specific level, AUTALIC provides an important recent example by introducing a contextual dataset for anti-autistic ableist language \cite{rizvi2025autalic}. Nevertheless, condition-specific computational resources for ableist harm remain scarce, and the linguistic and social forms of stigma can vary substantially across different disabilities and communication conditions. These limitations make it difficult to assume that models or annotation schemes developed for general toxicity, or even for one disability context, can be directly transferred to another condition.

Stuttering is a clear example of the broader challenge of distinguishing harmful disability-related language from non-harmful uses of the same expressions. Previous social-media studies on stuttering have documented many forms of discourse, including negative representations, metaphorical uses, ridicule, awareness, advocacy, and support \cite{azios2022twitter,rasoli2025stuttering}. However, these studies have mainly described how stuttering is represented and discussed online. They have focused on content, sentiment, engagement, and lived experience rather than building supervised datasets and models that distinguish among stigmatizing, non-stigmatizing/contextual, and supportive discourse. To the best of our knowledge, no publicly available dataset has been created specifically for this three-class task, and no target-aware computational framework has been developed for detecting stuttering-related stigma. This gap matters because stuttering-related stigma is not always expressed through direct mockery or insult. It may also appear as written imitation, assumptions of incompetence, devaluation, or exclusion from speaking and other communication-related roles.

No published computational model or benchmark is currently available for Turkish stuttering-related stigma detection, largely because no task-specific dataset has been available for this problem. Existing Turkish harmful-language datasets are useful for general offensive-language detection, but they were not designed to identify disability-related or stuttering-specific stigma \cite{coltekin2020corpus,zampieri2020semeval}. Turkish also has substantial morphological variation, and informal social-media writing often includes elongated words, fragmented spellings, repeated punctuation, laughter markers, and other non-standard forms that can be used to represent stuttered speech. More importantly, expressions such as \textit{kekeme} (person who stutters), \textit{kekemelik} (stuttering), and forms derived from \textit{kekele-} or \textit{kekel-} (to stutter) can appear in harmful, neutral, self-referential, reported, clinical, or supportive posts. Therefore, the presence of these expressions shows that a post is about stuttering, but the surrounding context determines whether the post is stigmatizing, non-stigmatizing/contextual, or supportive.

To address these gaps and challenges, we introduce \textit{StutterStigma-TR}, a condition-specific Turkish dataset for stuttering-related stigma detection, and propose \textit{StutterStigma-BERT}, a target-aware modelling framework built on BERTurk \cite{schweter_2020_berturk}. StutterStigma-TR distinguishes among \textit{Stigmatizing},
\textit{Non-stigmatizing/contextual}, and \textit{Supportive} discourse while
retaining stuttering-related target expressions to support target-aware
modelling. Building on prior work showing that disability-related language can be misinterpreted by automated systems and that ableist harm may depend on context \cite{hutchinson2020social,venkit_etal_2023,phutane2025cold}, this design directly addresses the need to distinguish harmful meaning from the mere presence of disability-related language. Rather than treating stuttering-related vocabulary as direct evidence of harm, \textit{StutterStigma-BERT} integrates representations of the marked target with its surrounding context, character-level and orthographic information, and discourse-related cues. This target-aware design is intended to help distinguish neutral, self-referential, informational, and supportive uses of stuttering-related language from stigmatizing uses, including subtler forms of stigma that depend on contextual meaning.

The primary task is formulated as a three-class discourse classification problem that distinguishes among \textit{Stigmatizing}, \textit{Non-stigmatizing/contextual}, and \textit{Supportive} discourse. This formulation captures not only whether stigma is present, but also whether non-stigmatizing stuttering-related content is contextual or explicitly supportive. To examine stigma detection as a simpler applied decision problem, we also define a secondary binary task by merging the \textit{Non-stigmatizing/contextual} and \textit{Supportive} classes into a single \textit{Non-stigma} category. The resulting task distinguishes \textit{Stigmatizing} from \textit{Non-stigma} discourse.

The proposed modelling framework is evaluated through controlled 
comparisons spanning lexical and contextual baselines, a target-marked 
BERTurk reference configuration, and target-aware hybrid models. For the 
primary three-class task, the experiments examine how performance changes 
across configurations that incorporate explicit target representation, target-
conditioned context, orthographic and character-level information, and discourse-
oriented cues. These comparisons allow us to evaluate whether richer target-
aware configurations improve on simpler reference models, while components 
introduced within combined configurations are interpreted at the configuration 
level rather than as fully independent effects. For the secondary binary task, 
validation-based decision-threshold selection is evaluated separately by 
selecting the threshold that maximizes validation Macro-F1 and applying it 
unchanged to the held-out test set. 
The main contributions of this study are as follows:

\begin{itemize}

\item 
We introduce StutterStigma-TR, a condition-specific Turkish social-media
dataset containing 6,019 X/Twitter posts annotated as
\textit{Stigmatizing}, \textit{Non-stigmatizing/ contextual}, or
\textit{Supportive}. The dataset retains stuttering-related target expressions
for target-aware modelling and provides a new resource for
studying stuttering-related stigma in Turkish.

\item 
We propose StutterStigma-BERT, a target-aware framework designed
specifically for Turkish stuttering-related stigma detection. The proposed configuration extends the BERTurk model by integrating global post information,
explicit target-span representation, target-conditioned context, character-level
and handcrafted orthographic information, and gated global and target-local
discourse cues to improve the performance.

\item 
The StutterStigma-BERT framework is proposed, optimized, and evaluated for both three-class and binary settings. 
Across 20 random seeds, the final three-class model reaches a mean Macro-F1 of
73.54\%, and the final binary model reaches 85.17\% using
validation-selected decision thresholds, under the matched three-seed protocol, both obtain higher Macro-F1 than plain BERTurk.

\item 
We provide a comprehensive evaluation of the proposed StutterStigma-BERT framework through controlled three-seed ablation studies of its individual components, training-diagnostic analyses, final 20-seed robustness experiments, and binary decision-threshold sensitivity analysis, with model and configuration selection based exclusively on internal-validation Macro-F1 rather than held-out test performance. We further assess backbone dependence by replacing BERTurk with Turkish Mamba and TurkishBERTweet under matched configurations for the primary three-class task, demonstrating that the proposed BERTurk-based configuration consistently achieves higher mean internal-validation Macro-F1 scores than the corresponding alternative-backbone implementations.

\item 
We provide a comprehensive comparative evaluation of the proposed framework against a diverse set of existing approaches, including three lexical and contextual baselines and four task-adapted neural models trained under a common protocol, three off-the-shelf harmful-language classifiers evaluated through direct transfer, and two zero-shot instruction-tuned LLMs, in which the proposed framework obtains the highest mean Macro-F1 in each comparison setting.

\end{itemize}

Taken together, this study conceptualizes stuttering-related stigma detection as a disability-informed classification task in which predictions require contextual interpretation. The proposed framework is intended to support research, discourse auditing, and human review of Turkish-language X/Twitter content, rather than serve as a fully automated content-moderation system. 

The remainder of the article is organized as follows.
Section~\ref{sec:related_work} establishes the conceptual and computational
background of the study. Section~\ref{sec:dataset} describes the construction,
annotation, and composition of the StutterStigma-TR dataset.
Section~\ref{sec:methodology} presents the task formulation, preprocessing
procedures, and proposed StutterStigma-BERT framework.
Section~\ref{sec:experiments_results} describes the experimental setup and
evaluation metrics and reports the training diagnostics, configuration and
backbone analyses, final robustness results, threshold analysis, broader
model comparisons, zero-shot LLM evaluation, and repeated-run error analysis. Section~VI discusses the findings and limitations, and Section~VII concludes the paper.

\section{Related Work}
\label{sec:related_work}
This section reviews the conceptual and computational background of the study. It first discusses stuttering as a multidimensional communication experience and explains why stigma and ableism are relevant to stuttering-related discourse. It then reviews previous work on stuttering and social media, with attention to the coexistence of stigmatizing, contextual, personal, and supportive uses of stuttering-related language. The section then reviews existing datasets and the Turkish stuttering resource gap, examines disability bias and ableist harm in NLP, and finally discusses target-aware and orthography-aware modelling approaches relevant to the proposed framework. Together, these strands provide the conceptual and methodological basis for distinguishing the contribution of the present study from existing disability-related and target-dependent NLP research.

\subsection{Stuttering, Stigma, and Ableism Definitions}

Stuttering is increasingly understood as a multidimensional communication experience rather than only an observable disruption in speech fluency. Although repetitions, prolongations, and blocks are central to the clinical description of stuttering, people who stutter may also experience anticipation, avoidance, concealment, negative listener reactions, reduced participation, and social evaluation \cite{asha_fluency_disorders,tichenor2019stuttering,tichenor_etal_2022}. This perspective is central to stigma research because the social consequences of stuttering are shaped not only by speech characteristics but also by how stuttered speech is interpreted and evaluated by others. Public attitudes toward stuttering often involve inaccurate assumptions about personality, confidence, competence, intelligence, employability, and social desirability \cite{boyle2013assessment,ozdemir2011public,louis2011public,st_louis_2015,gabel2004measuring,gerlach2018stuttering}. Such stereotypes extend judgments about speech fluency to evaluations of a person’s capability, credibility, and suitability for communication-demanding social or professional roles \cite{gabel2004measuring,gerlach2018stuttering}. Stuttering-related stigma can therefore take the form of direct ridicule, but it can also operate through competence-based judgments concerning credibility and communicative suitability.

Stigma theory helps explain how a speech difference acquires broader social consequences. Goffman \cite{goffman2009stigma} conceptualized stigma as a socially discrediting relation between an attribute and its social context, whereas Link and Phelan \cite{link2001conceptualizing} emphasized the interconnected processes of labelling, stereotyping, separation, status loss, and discrimination. Applied to stuttering, these frameworks help explain how stuttered speech may be interpreted as evidence of nervousness, weakness, incompetence, or communicative failure \cite{boyle2013assessment,st_louis_2015,gabel2004measuring, gerlach2018stuttering, ozdemir2011public, louis2011public}. 

Ableism provides a complementary account of how particular bodily and functional norms become privileged \cite{campbell2009contours, goodley2014dis}. Recent research on stuttering has increasingly drawn attention to how social and clinical environments may privilege fluent speech as a normative standard and evaluate stuttered speech against dominant expectations of communication \cite{constantino2022stuttering, gerlach2022interrupting, tichenor2022point}. 

Together, stigma and ableism distinguish the mere reference to stuttering from the social evaluation constructed around that reference. Accordingly, the same stuttering-related expression may contribute to stigmatizing, contextual, or supportive discourse depending on its stance, framing, and communicative function, providing the conceptual basis for the target-dependent formulation adopted in this study.

\subsection{Stuttering-Related Discourse on Social Media}

Social media is a consequential setting for stuttering-related discourse because the same platforms that facilitate self-expression, information exchange, community connection, and advocacy may also circulate misinformation, mockery, and negative stereotypes \cite{fuse2016impact,bidakhavidi2025social}. They may also create or intensify opportunities for cyberbullying and other harmful interactions involving people who stutter \cite{ccavdar2026stuttering,nicolai2018retrospective}. This coexistence of supportive and harmful functions makes online stuttering discourse a heterogeneous communicative environment rather than a uniformly positive or negative domain.

Previous studies indicate that stuttering-related content varies across platforms, media formats, sources, and communicative purposes. Video-based analyses have identified educational, entertainment-oriented, self-disclosing, and misleading content, while reliability, quality, and audience responses may vary by source type and platform \cite{ccavdar2025reliable, donaher2014portrayal}. On X/Twitter, stuttering has been used metaphorically to denigrate others or imply incompetence, yet positive, neutral, and negative representations also coexist in samples of highly engaged posts \cite{azios2022twitter, rasoli2025stuttering}. Together, these findings show that online representations of stuttering differ not only in tone but also in source, purpose, and stance.

Beyond its representational content, online stuttering discourse may also have consequences for participation and perceived safety. Cyberbullying experiences have been associated with adverse psychosocial outcomes among people who stutter \cite{nicolai2018retrospective}, while qualitative evidence suggests that anticipated negative reactions may encourage more cautious self-disclosure, greater preparation before posting, privacy-seeking, and preferences for safer or more passive forms of interaction \cite{ccavdar2026stuttering}. The significance of online stigma therefore extends beyond the content of individual posts to the conditions under which people who stutter participate, disclose personal experiences, and remain visible online.

For automated classification, the central challenge is therefore not lexical relevance but discourse interpretation. Posts containing the same stuttering-related expression may enact stigma, describe personal or clinical experience, report or quote discrimination, or oppose ableist language. Consequently, explicitly negative vocabulary may occur in supportive counter-speech, whereas indirect stigma may be expressed without overtly harmful terms. This asymmetry makes keyword-based classification unreliable and motivates models that interpret the marked target in relation to its surrounding context and stance.

\subsection{Existing Datasets and the Turkish Stuttering Resource Gap}

Annotated datasets and benchmark resources are central to the development and evaluation of NLP systems, but their availability remains uneven across languages. Widely studied languages such as English are represented across a broader range of tasks and benchmark collections, whereas many other languages have more limited task-specific resources \cite{joshi2020state}. This imbalance can restrict the development, comparison, and reproducible evaluation of computational models for specialized language-processing tasks.

Turkish has an established and growing NLP resource base, but dataset coverage remains uneven across domains and specialized tasks \cite{coltekin2023resources}. This limitation is particularly relevant to socially sensitive and condition-specific applications. Existing Turkish harmful-language resources have mainly focused on general offensive-language detection. For example, Çöltekin \cite{coltekin2020corpus} introduced a Turkish social-media corpus for offensive-language identification, while OffensEval 2020 included Turkish within a multilingual offensive-language benchmark \cite{zampieri2020semeval}. Although these resources provide important benchmarks for harmful-language research in Turkish, their annotation schemes were not designed to distinguish disability-related or condition-specific stigma from contextual and supportive discourse.

Condition-specific disability datasets remain comparatively scarce. AUTALIC provides a recent example of a context-rich dataset developed specifically for anti-autistic ableist language \cite{rizvi2025autalic}. In contrast, previous stuttering-related social-media studies have primarily examined representation, content, sentiment, engagement, and lived experience rather than providing supervised datasets for stigma classification \cite{azios2022twitter,rasoli2025stuttering}. To the best of our knowledge, no publicly available Turkish dataset has been developed specifically to distinguish stigmatizing, non-stigmatizing/contextual, and supportive stuttering-related discourse. This resource gap limits reproducible benchmarking and supervised computational research on Turkish stuttering-related stigma. StutterStigma-TR addresses this gap by introducing a condition-specific dataset of 6,019 annotated Turkish social-media posts with retained stuttering-related target expressions for target-aware modelling; the dataset will be publicly available upon acceptance of the paper.

\begin{table*}[t]
\caption{Representative Target-Aware and Orthography-Aware Computational Approaches in the Literature}
\label{tab:related_computational_approaches}
\centering
\footnotesize
\setlength{\tabcolsep}{4pt}
\begin{tabular}{p{2.7cm} p{2.8cm} p{4.0cm} p{6.5cm}}
\hline
\textbf{Study} & \textbf{Task} & \textbf{Main approach} & \textbf{Summary of the Work} \\
\hline

Tang et al.\ \cite{tang_etal_2016}
& Target-dependent sentiment
& Target-dependent LSTM
& Provides a foundation for representing a specified target in relation to its surrounding context. \\

Ma et al.\ \cite{ma2017interactive}
& Aspect-level sentiment
& Interactive attention network
& Demonstrates attention-based modelling of interactions between an aspect representation and its context. \\

Sun et al.\ \cite{sun_etal_2019}
& Aspect-based sentiment
& BERT with aspect-derived auxiliary sentences
& Illustrates the incorporation of explicit aspect information into BERT-based contextual classification. \\

Wu and Ong \cite{wu_ong_2021}
& Targeted aspect-based sentiment
& Context-guided BERT
& Provides a BERT-based approach for jointly representing target information and relevant contextual evidence. \\

Mutlu and Özgür \cite{mutlu2022dataset}
& Turkish targeted sentiment
& BERTurk with explicit target markers and target-output variants
& Provides a Turkish example of incorporating explicit target information and target-related token representations into BERTurk-based classification. \\

Zhang et al.\ \cite{zhang_etal_2015_charcnn}
& General text classification
& Character-level CNN
& Provides a character-level modelling basis for representing spelling and surface-form variation. \\

\textbf{Ours}
& \textbf{Turkish stuttering-related stigma}
& \textbf{StutterStigma-BERT}
& \textbf{Integrates explicit stuttering-target representation, target-conditioned within-post context, character-level and orthographic information, and gated global and target-local discourse cues.} \\

\hline
\end{tabular}
\end{table*}

\subsection{Disability Bias and Ableist Harm in NLP}

Automated harmful-language detection has largely been organized around broad labels such as hate speech, offensive language, toxicity, harassment, and cyberbullying. These frameworks provide useful methodological foundations, but their label definitions are not necessarily suited to condition-specific ableist harm. Disability-related harm may be expressed through stereotyping, devaluation, or exclusion that is not adequately captured by conventional markers of overt abuse. Disability-related NLP therefore requires task definitions that reflect the social meanings and stereotypes associated with specific disability-related targets and experiences.

Research on disability-related NLP highlights two main risks. First, sentiment and toxicity models may assign more negative or toxic scores to otherwise neutral disability-related language, creating false positives for self-description, educational content, and advocacy \cite{hutchinson2020social,venkit_etal_2023}. Second, ableist harm can be subtle and context-dependent rather than explicitly insulting \cite{heung2022nothing}. In addition, AI judgments of ableist harm do not always align with those of disabled participants \cite{phutane2025cold}, and recent evaluations show that language models may rely on surface-level cues and miss harmful meanings that depend on context \cite{rizvi2025beyond}. These findings show that disability-related harm cannot be reliably identified from surface negativity alone. Recent condition-specific research provides a more direct basis for context-sensitive ableism detection. Rizvi et al.\cite{rizvi2025autalic} introduced AUTALIC, a context-rich dataset designed to support the identification of anti-autistic ableist language while preserving the surrounding discourse. Related evaluation work showed that large language models could recognize autism-related terminology but often relied on surface-level cues and missed harmful meanings that depended on context, speaker identity, and potential impact \cite{rizvi2025beyond}. Together, these studies demonstrate the importance of condition-specific annotation and context-preserving evaluation for disability-related harm detection.

Taken together, existing approaches leave important limitations for the present problem: general sentiment and toxicity systems may misclassify disability-related language, while existing condition-specific ableism resources focus on other disability contexts. We address this gap with a target-aware classifier that interprets the marked stuttering-related expression in relation to its context rather than treating it as harmful by default.

\subsection{Target-Aware and Orthography-Aware Approaches}

Although the presented task in this work is not a sentiment-analysis task, it shares an important structural property with target-dependent and aspect-based classification: the interpretation of a text depends on a specific target and its surrounding context. Tang et al. \cite{tang_etal_2016} introduced target-dependent models that represented a specified target together with its contextual information, while Ma et al. \cite{ma2017interactive} used interactive attention to model the relationship between an aspect and its context. These studies established the importance of representing the target--context relationship rather than relying only on a global text representation.

Later BERT-based approaches incorporated target or aspect information directly into contextual language representations. Sun et al. \cite{sun_etal_2019} used BERT with aspect-related auxiliary sentences for aspect-based sentiment classification, while Wu and Ong \cite{wu_ong_2021} proposed a context-guided BERT approach for targeted aspect-based sentiment analysis. In Turkish, Mutlu and Özgür \cite{mutlu2022dataset} developed BERTurk-based models for targeted sentiment analysis using explicit target markers and target-related output representations. Their work provides a particularly relevant example of incorporating explicit target information into Turkish BERT-based classification.

Orthographic and character-level information provides a complementary modelling direction for informal social-media text. Turkish has productive agglutinative morphology, allowing lexical roots to occur in many inflected and derived forms \cite{oflazer1994twolevel,oflazerKuruoz1994tagging}. Informal social-media writing also contains substantial non-standard and paralinguistic variation in spelling and written expression \cite{eisenstein2013badlanguage,pavalanathanEisenstein2016emojis}. Character-level convolutional models provide a way to represent such spelling and surface-form variation directly from character sequences \cite{zhang_etal_2015_charcnn}. Table~\ref{tab:related_computational_approaches} summarizes representative target-aware and orthography-aware approaches from the literature and shows how they relate to the present framework.

Despite these advances, the reviewed approaches were developed primarily for sentiment analysis, aspect-based classification, or general text classification rather than condition-specific stigma detection. 
Our proposed model, \textit{StutterStigma-BERT}, addresses this methodological gap by adapting target-aware and orthography-aware modelling principles to classify Turkish stuttering-related expressions according to their surrounding context and discourse function.

\section{Proposed StutterStigma-TR Dataset}
\label{sec:dataset}

\begin{figure}[htbp]
\centering
\includegraphics[width=0.45\linewidth]{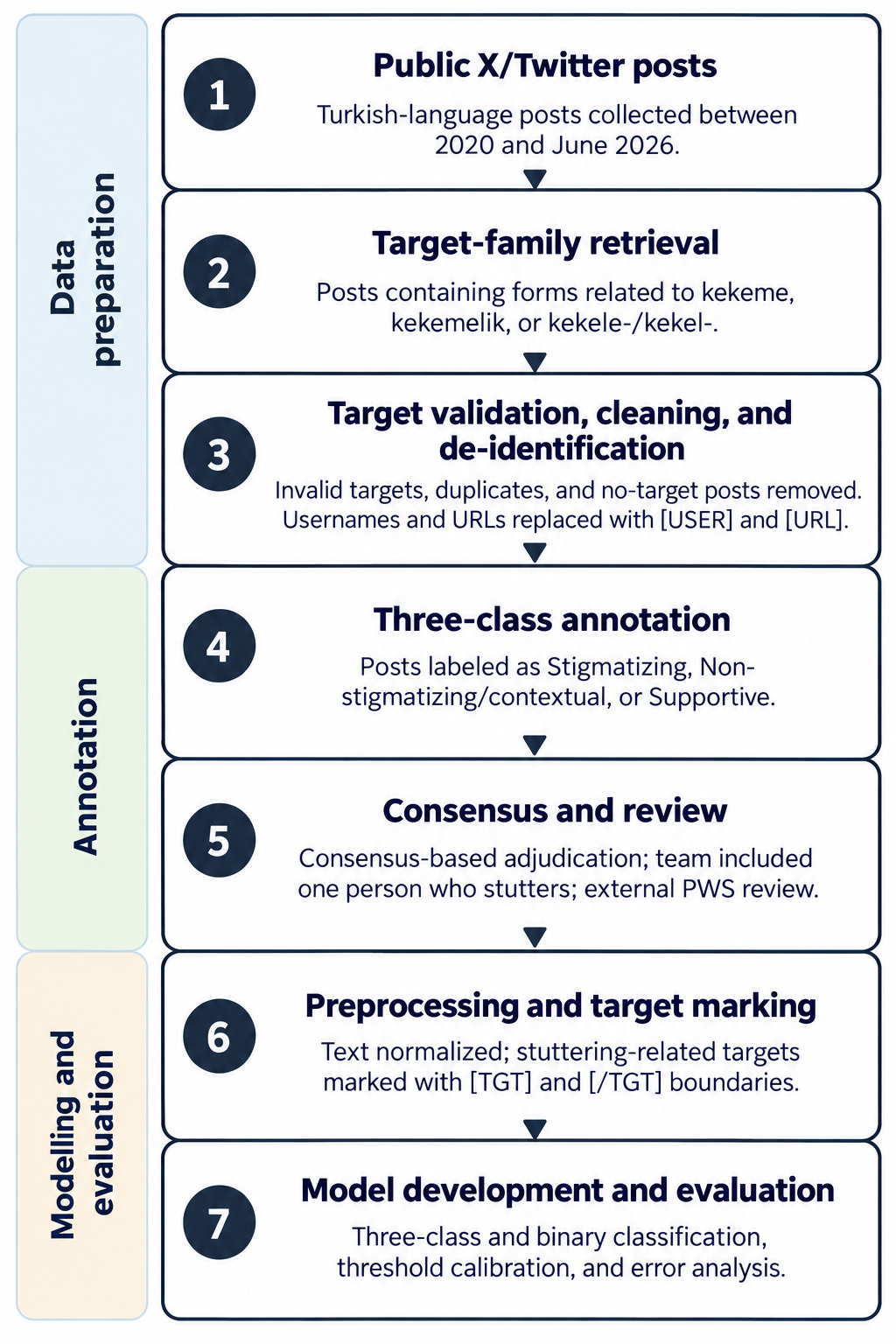}
\caption{Main workflow of the study, from Turkish stuttering-related data
collection and annotation to modelling and evaluation.}
\label{fig:study_workflow}
\end{figure}

As illustrated in the main workflow of this study in Fig.~\ref{fig:study_workflow}, the first stage involves the development of a novel dataset, including raw data collection from X/Twitter and subsequent annotation, which provides the foundation for developing the proposed detection model. This section describes the construction of \emph{StutterStigma-TR} dataset, including data collection and screening, the annotation scheme, annotator consensus and lived-experience evaluation, and the composition of the final model-ready dataset. 

\subsection{Proposed Dataset Construction: StutterStigma-TR}

\textbf{StutterStigma-TR dataset}\footnote{To request access to the StutterStigma-TR dataset, please contact the authors of the paper.} is designed as a condition-specific
Turkish social-media dataset for detecting stuttering-related stigma
and analysing related discourse patterns. The dataset consists of public Turkish-language posts collected from X/Twitter between January 2020 and June 2026. It focuses on posts that contain explicit Turkish expressions referring to stuttering, people who stutter, or stuttering-like speech production. Retrieval covers three predefined lexical target families: \textit{kekeme}, including forms referring to a person who stutters; \textit{kekemelik}, referring to stuttering as a condition or concept; and verbal forms derived from \textit{kekele-} or \textit{kekel-}, referring to stuttering or stuttering-like speech production. Because Turkish language permits productive inflectional and derivational extensions, the retrieval and screening procedures accommodate multiple surface forms derived from these target families.

Target-family matching serves only as an initial retrieval criterion and does not determine the class label. The dataset contains 6,262 posts after manual collection and initial annotation. After annotation, these records undergo quality-control filtering to construct the model-ready dataset. A post is retained if it contains a valid stuttering-related target that can be reliably identified and if the available text provides sufficient context for interpretation. Posts are excluded at this stage if they are duplicates, if no valid stuttering-related target remains after inspection, if an apparent target expression does not actually refer to stuttering, or if the relevant target span cannot be identified reliably.

Duplicate removal is performed before the train--test split. To identify duplicates, a temporary comparison version of each post is created by applying Unicode NFKC normalization, replacing usernames and URLs with standardized tokens, collapsing whitespace, and ignoring letter case. Posts that become identical after these operations are treated as duplicates, and one record is retained. This procedure captures exact duplicates and simple surface variants rather than broader paraphrases. Letter case is ignored only during duplicate detection; the original casing is preserved in the model input because capitalization can provide potentially relevant orthographic information. Across these post-annotation quality-control procedures, 243 posts are excluded, reducing the dataset from 6,262 to 6,019 posts. The final model-ready dataset retains the final consensus label, the detected target span and root-family information, and the text representations required for model training and evaluation. As part of dataset preparation and privacy protection, usernames and URLs are
replaced with the standardized tokens \texttt{[USER]} and \texttt{[URL]},
respectively. The original usernames and URLs are not retained in the
de-identified text used for subsequent model development. This de-identified
representation forms the basis of the model-ready StutterStigma-TR dataset
and the dataset materials prepared for public release.

The final model-ready dataset is partitioned once into training and held-out test sets using an 80/20 split stratified by the three-class labels. The training partition contains 4,815 posts, and the held-out test partition contains 1,204 posts. This fixed partition is reused across all model families to ensure that they are evaluated on the same test examples for fair comparison. 
Within each random-seed run, the training partition is further divided into an internal training subset and a stratified validation subset corresponding to 10\% of the training partition, while the held-out test partition remains unchanged. The final three-class distribution is summarized in Table~\ref{tab:dataset_distribution}. For the secondary binary task, the same partition is retained and only the label mapping is changed. The \textit{Stigmatizing} class contains 2,492 posts, including 1,994 training and 498 test posts. Merging the \textit{Non-stigmatizing/contextual} and \textit{Supportive} classes produces a \textit{Non-stigma} class of 3,527 posts, including 2,821 training and 706 test posts.

\begin{table}[t]
\caption{Distribution of the Final Model-Ready StutterStigma-TR Dataset}
\label{tab:dataset_distribution}
\centering
\small
\setlength{\tabcolsep}{4pt}
\renewcommand{\arraystretch}{1.08}

\begin{tabular}{@{}p{0.47\columnwidth}rrr@{}}
\toprule
\textbf{Label} &
\textbf{Total} &
\textbf{Training} &
\textbf{Test} \\
\midrule

Stigmatizing &
2,492 &
1,994 &
498 \\

Non-stigmatizing/contextual &
1,803 &
1,442 &
361 \\

Supportive &
1,724 &
1,379 &
345 \\

\midrule
Total &
6,019 &
4,815 &
1,204 \\

\bottomrule
\end{tabular}
\end{table}

Each post is also assigned a target-family category based on the stuttering-related expressions it contains. Posts containing expressions from only one family are categorized as \textit{kekeme}, \textit{kekemelik}, or \textit{kekele-/kekel-}, whereas posts containing expressions from more than one family are categorized as \textit{Mixed}. Most posts are assigned to the \textit{kekeme} category ($n=4,178$, 69.4\%), followed by \textit{kekemelik} ($n=1,012$, 16.8\%), \textit{Mixed} ($n=569$, 9.5\%), and \textit{kekele-/kekel-} ($n=260$, 4.3\%). The target-family category itself is not used as a prediction label, model input feature, or model-selection criterion, but it is retained for descriptive reporting and error interpretation. 
The collected dataset is subsequently preprocessed to prepare it as input
for the proposed model, as described in Section~\ref{sec:preprocessing}.

\subsection{Annotation Scheme}

Each post receives one post-level label based on how the marked stuttering-related target expression or expressions function within the available post text. The primary annotation scheme comprises three mutually exclusive labels: \textit{Stigmatizing}, \textit{Non-stigmatizing/contextual}, and \textit{Supportive}. A post is labeled \textit{Stigmatizing} when it expresses, endorses, normalizes, or reproduces without clear distancing a harmful framing of stuttering, people who stutter, or stuttering-like speech. This includes direct ridicule, written imitation or parody, negative stereotyping, the use of stuttering as an insult or shorthand for incompetence, devaluation, and exclusion from speaking or other communication-demanding roles. A post is labeled \textit{Non-stigmatizing/contextual} when it refers to stuttering, a person who stutters, or stuttering-like speech without expressing a clear stigmatizing or supportive stance. This category includes neutral, descriptive, clinical, informational, personal, and situational uses in which the available post text does not provide sufficient evidence of either harmful framing or explicit support. A post is labeled \textit{Supportive} when it expresses an explicitly supportive, awareness-oriented, or anti-stigma stance toward stuttering or people who stutter. This category includes posts that defend people who stutter, reject or criticize mockery and discrimination, challenge negative stereotypes, express solidarity, promote acceptance and inclusion, or affirm stuttering as a valid form of communication.

Reported or quoted stigma is coded according to the post's stance toward the harmful content rather than the presence of stigmatizing wording alone. A post is labeled \textit{Stigmatizing} when it endorses, normalizes, or reproduces the harmful framing without clear distancing. It is labeled \textit{Supportive} when it explicitly rejects, criticizes, or challenges that framing. A post is labeled \textit{Non-stigmatizing/contextual} when it primarily reports or quotes the harmful content and the available text does not establish either endorsement or opposition. Posts labeled \textit{Stigmatizing} are additionally assigned one dominant stigma subtype: mockery/belittlement, imitation/parody, attribution of incompetence, devaluation, or exclusion/silencing. When a post contains more than one stigmatizing pattern, the subtype that best represents its dominant harmful function is assigned. These subtype labels are retained for descriptive analysis and error interpretation and are not used as main labels or supervised prediction targets in either the primary three-class task or the secondary binary task.

\subsection{Annotators, Consensus, and Lived-Experience Review}

Annotation is conducted by four annotators, all of whom have both clinical and academic experience related to stuttering. The team comprises two doctoral students who also hold lecturer positions, one master's student, and one master's graduate in speech and language therapy. The master's graduate is also a person who stutters. Before annotation, the research team develops written guidelines for the three main labels. The guidelines define each class, describe how to interpret the marked target expression or expressions within the available post text, and specify decision rules for ambiguous cases. Annotators are instructed to evaluate the post-level stance, particularly in cases involving reported or quoted stigma, irony, sarcasm, and anti-stigma responses.

Under the standard annotation procedure, the dataset is distributed among the four annotators, and each post is initially labelled by one assigned annotator using the shared annotation guidelines. Annotators flag examples that they regard as ambiguous, context-dependent, or otherwise difficult. These cases are reviewed during repeated consensus meetings, in which the team jointly considers the identified target expression, the surrounding post text, the post-level stance, and competing label interpretations.

A separate randomly selected common subset of 645 posts is independently
labelled by all four annotators for inter-annotator reliability analysis. Only the main three-class annotation scheme is used for this reliability calculation. The independent labels are recorded before consensus discussion, and disagreements within this common subset are reviewed subsequently so that an adjudicated label can be retained in the final dataset. Consensus review is not restricted to the 645-post reliability subset; difficult or disputed cases encountered during the wider annotation process are also discussed.

Within the 645-post reliability subset, full agreement is defined as all four
annotators assigning the same main label (4/4), partial agreement as a
three-of-four majority (3/4), and insufficient agreement as the absence of a
three-annotator majority (2--2 or 2--1--1). Full agreement is observed for
427 posts (66.2\%), partial agreement for 178 posts (27.6\%), and insufficient
agreement for 40 posts (6.2\%) before adjudication. Fleiss' kappa for the
three-class annotation is $\kappa = 0.66$ \cite{fleiss1971measuring}.

\begin{figure*}[!t]
\centering
\includegraphics[width=1\textwidth]
{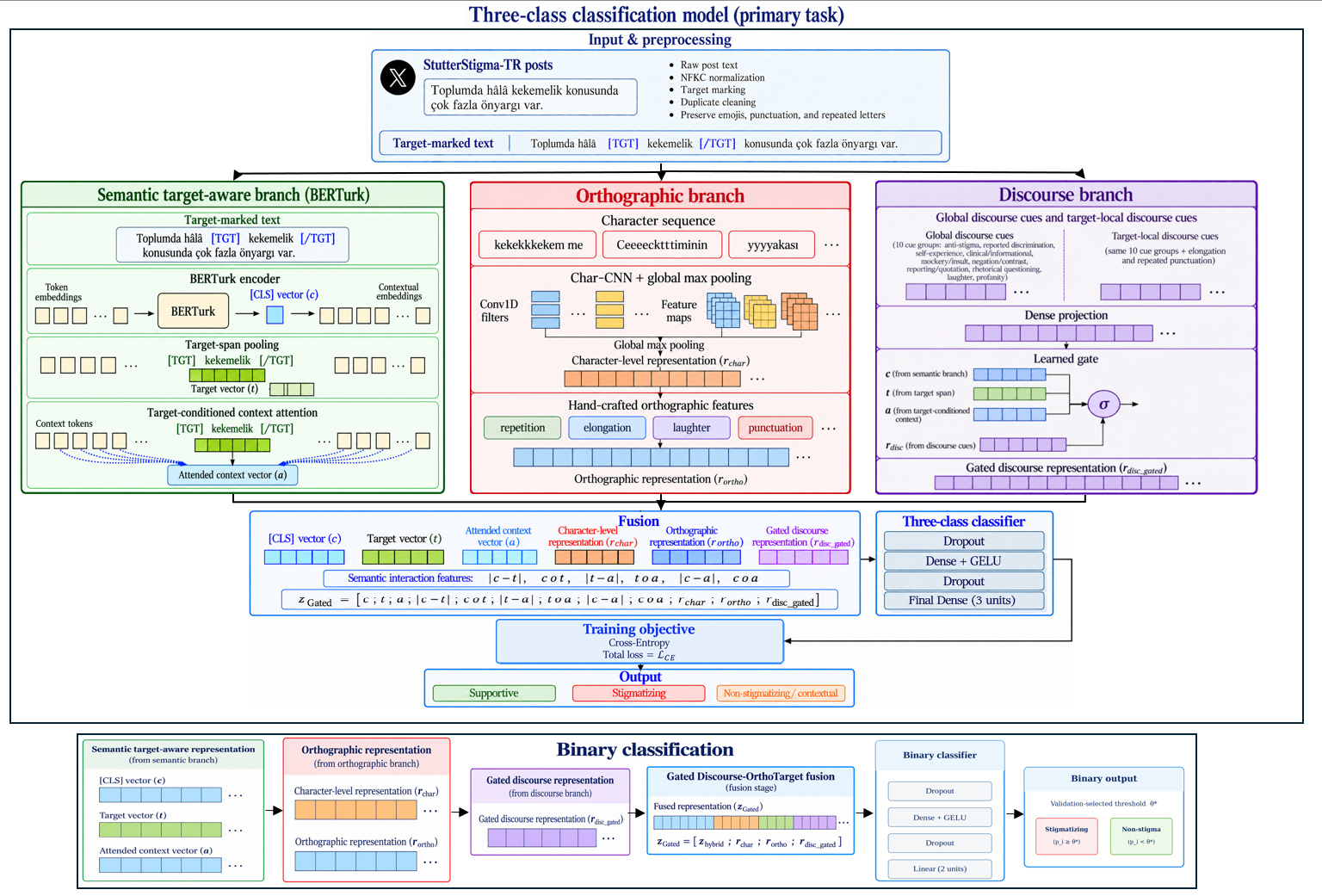}
\caption{
Overall architecture of the proposed StutterStigma-BERT framework for the
primary three-class and secondary binary classification tasks. Both
configurations integrate semantic target-aware, orthographic, and gated
discourse representations. The three-class pathway uses a three-unit
classifier, whereas the binary pathway uses a two-unit classifier followed
by validation-selected decision-thresholding.
}
\label{fig:gated_architecture}
\end{figure*}

An independent lived-experience review is conducted by an adult person who stutters and is not involved in the primary annotation or consensus process. The external evaluator independently labels a randomly selected subset of
750 posts using the same three-class annotation scheme, and these labels are
compared with the final consensus labels. Agreement is observed for 628 of the 750 posts, corresponding to 83.7\% observed agreement. Cohen's kappa for agreement between the external evaluator and the consensus labels is $\kappa = 0.62$ \cite{cohen1960coefficient}. The inter-annotator reliability analysis and the independent lived-experience review provide complementary evidence about agreement in the annotation scheme and highlight recurrent areas of disagreement. These areas include distinctions between neutral or contextual mention and supportive counter-speech, reported and reproduced stigma, and subtle stigmatizing meanings such as attribution of incompetence. These observations are subsequently considered in the error analysis. 
The de-identified dataset representation established during dataset preparation
is retained throughout model development and reporting. The article reports
aggregate statistics, model results, and error patterns without reproducing
identifiable user account information.

\section{Proposed Model and Methodology}
\label{sec:methodology}
This section describes the study design and task formulation after dataset collection, followed by text preprocessing and target marking and the proposed StutterStigma-BERT framework. We describe how BERTurk \cite{schweter_2020_berturk} is extended with semantic target-aware, orthographic, and discourse-aware representations for the three-class and binary classification schemes. The same preprocessing and target-aware input-construction procedures are applied consistently to the training, validation, and held-out test samples of our dataset.

\subsection{Three-Class vs. Binary Classification Tasks in this Study}

This study uses a supervised post-level text classification design to develop and evaluate models for detecting stuttering-related stigma in public Turkish-language X/Twitter posts. The unit of analysis is an individual post containing one or more stuttering-related target expressions identified by the rule-based target-detection pipeline. The objective is to assign a post-level label according to how the marked target expression or expressions function within the surrounding discourse. As illustrated in Fig. \ref{fig:gated_architecture}, the framework is evaluated on two classification tasks: three-class and binary classification.

\begin{figure}[htbp]
\centering
\includegraphics[width=0.6\linewidth]{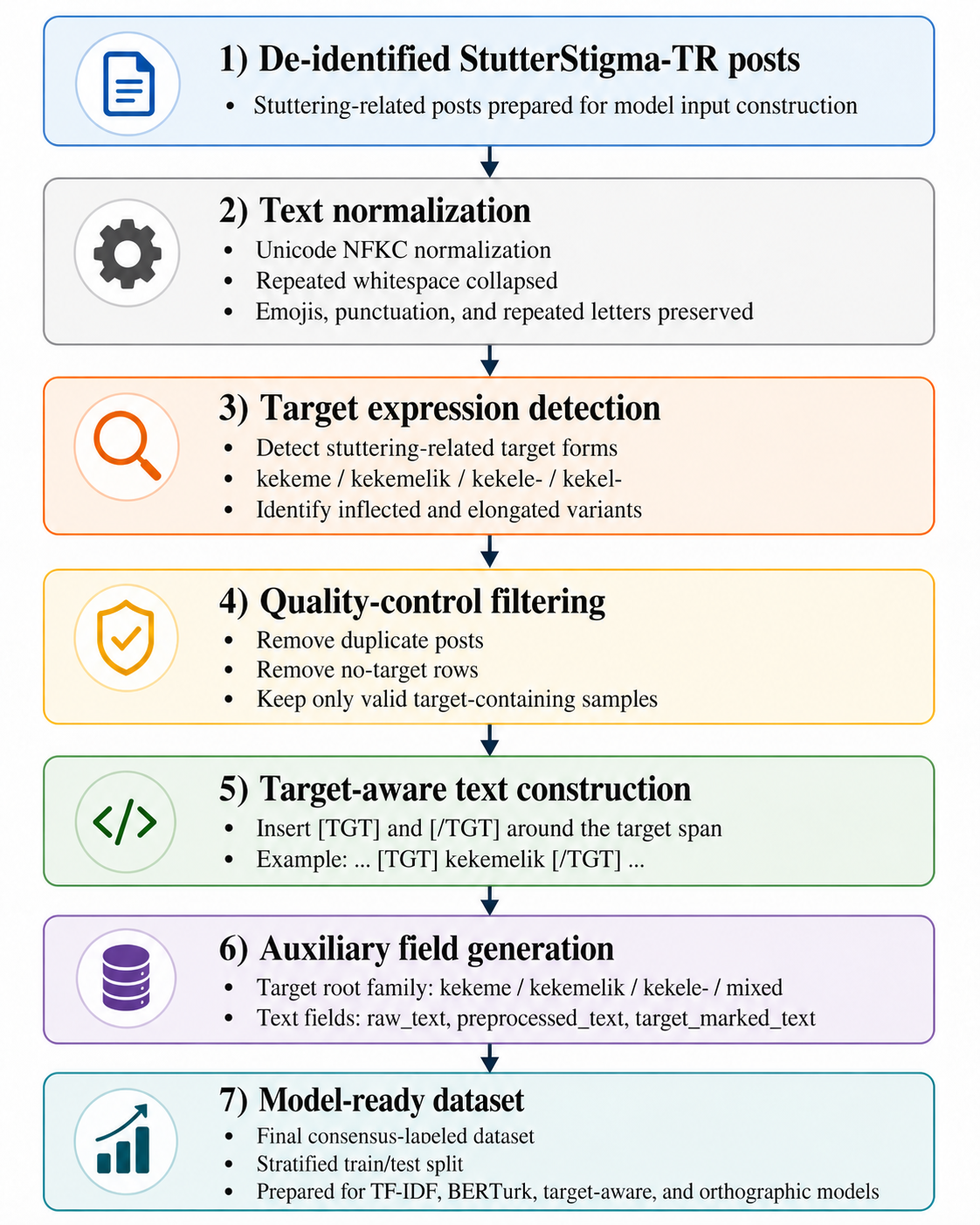}
\caption{Main pipeline for preprocessing the de-identified
\emph{StutterStigma-TR} dataset text and constructing target-aware inputs for the
proposed \emph{StutterStigma-BERT} model.}
\label{fig:target_input_pipeline}
\end{figure}

The primary task is formulated as a three-class classification problem. Each post is assigned one of three labels: \textit{Stigmatizing}, \textit{Non-stigmatizing/contextual}, or \textit{Supportive}. This formulation is selected because stuttering-related online discourse includes harmful, neutral/contextual, and supportive uses of similar target expressions. Keeping these categories separate allows the models to be evaluated on whether they can distinguish different discourse functions of stuttering-related language. A secondary binary task is also defined for an applied stigma-detection setting. In this task, \textit{Stigmatizing} posts are treated as the positive class, while \textit{Non-stigmatizing/contextual} and \textit{Supportive} posts are merged into a single \textit{Non-stigma} category. This binary formulation tests whether models can separate stigmatizing posts from all other stuttering-related posts.

Both tasks are treated as complementary rather than interchangeable. They use the same underlying posts and fixed train--test partition, with only the label mapping changed for the binary analysis. Within each task, all model families are evaluated on the same data partitions, enabling controlled comparisons among modelling strategies. Performance values are interpreted separately across the two tasks because their label spaces and decision problems differ.

\subsection{Text Preprocessing and Target Marking}
\label{sec:preprocessing}

Text preprocessing is performed before feeding the data into the model. The pipeline standardizes limited technical variation while preserving orthographic and paralinguistic cues that can convey emphasis, affect, or discourse stance. Accordingly, preprocessing is intentionally conservative. The main pipeline for preprocessing and constructing the target-aware data used as input to the proposed model is illustrated in Fig.~\ref{fig:target_input_pipeline}.
After dataset preparation, we apply text normalization to the de-identified
\emph{StutterStigma-TR} text. Unicode NFKC normalization is applied to reduce
inconsistencies in character representation, and repeated whitespace is
collapsed into a single space.

Additionally, emojis, punctuation, including repeated punctuation,
capitalization, hashtags, informal spelling, slang, profanity, laughter markers,
repeated letters, and imitation-like written forms are retained. Text is not
converted to lowercase because the BERTurk-based models use a cased encoder
and capitalization can convey emphasis or stance. Stopword removal, stemming,
and lemmatization are not applied because these operations can remove
information relevant to negation, reported speech, contrast, and supportive
stance.

In the next step, stuttering-related target spans are identified automatically using predefined lexical patterns for \textit{kekeme} (person who stutters), \textit{kekemelik} (stuttering), and forms derived from \textit{kekele-} or \textit{kekel-} (to stutter). The rule-based detection procedure accommodates inflectional and derivational extensions of these roots. For non-standard or elongated spellings, a fallback procedure compares repetition-compressed forms with the same target families. After detecting target expressions, a quality-control filtering step is applied. Duplicate posts and records without a valid stuttering-related target are removed, and only posts with reliably identifiable target spans are retained for subsequent target-aware input construction.

To construct target-aware text in the following step, each automatically detected target span is enclosed by the special boundary markers \texttt{[TGT]} and \texttt{[/TGT]}. For example, a post containing the expression \textit{kekemelik} is transformed so that the target appears as \texttt{[TGT] kekemelik [/TGT]} within the input sequence. These markers explicitly identify the automatically detected target span while preserving the surrounding post text. Both target detection and the insertion of the target boundary markers are performed automatically using the predefined rules. Formally, a target-marked input sequence is represented as:
\begin{equation}
\begin{array}{l}
X_{\mathrm{marked}} =
\langle \mathrm{[CLS]}, x_{1}, \ldots, x_{p}, \mathrm{[TGT]}, \\
\quad t_{1}, \ldots, t_{m}, \mathrm{[/TGT]}, x_{p+1}, \ldots, x_{n}, \mathrm{[SEP]} \rangle .
\end{array}
\label{eq:target_marked_input}
\end{equation}

where [CLS] denotes the classification token placed at the beginning of the
input sequence, [SEP] denotes the separator token marking the end of the
sequence, $t_1,\ldots,t_m$ denote the target-span tokens, and $x_i$ denotes
the surrounding non-target context tokens. The marker tokens themselves are not treated as part of the target span. If the target expression is split into multiple subword units, all tokens between \texttt{[TGT]} and \texttt{[/TGT]} are treated as target-span tokens. 
The target-boundary tokens \texttt{[TGT]} and \texttt{[/TGT]} are
introduced during preprocessing, whereas the dataset-level placeholders
\texttt{[USER]} and \texttt{[URL]} are introduced during dataset preparation.
All four tokens are included in the BERTurk tokenizer vocabulary. The
embedding matrix is resized accordingly so that these tokens can receive
trainable embeddings during fine-tuning. For target-aware models, the positions of \texttt{[TGT]} and \texttt{[/TGT]} are also used to identify the hidden states corresponding to the target span. After target-aware text construction, auxiliary fields are generated for each post, including the target-root family (\textit{kekeme}, \textit{kekemelik}, \textit{kekele-}/\textit{kekel-}, or Mixed) and the \texttt{raw\_text}, \texttt{preprocessed\_text}, and \texttt{target\_marked\_text} fields used by the downstream model components.

The preprocessing pipeline generates three model-specific text views. Preprocessed text without target markers is used for the TF-IDF baselines and Raw BERTurk, whereas target-marked text is used for the BERTurk-based target-aware configurations of our proposed \emph{StutterStigma-BERT} framework. 
The \texttt{raw\_text} representation is used for the character-level and
handcrafted orthographic streams, whereas \texttt{target\_marked\_text} is
additionally used to define the target-local context required by the discourse
stream.
The automatically detected target boundaries are additionally retained to construct the target-span masks required for pooling and target-conditioned attention. This design allows the target-aware models to represent the marked stuttering-related expression together with the context in which it functions \cite{tang_etal_2016,wu_ong_2021}.

\subsection{Proposed StutterStigma-BERT Framework}

The proposed StutterStigma-BERT framework is a BERTurk-based
target-aware architecture designed to model how a marked
stuttering-related target expression is used within its surrounding discourse.
As illustrated in Fig.~\ref{fig:gated_architecture}, the framework contains
three complementary representation branches: a semantic target-aware
branch, an orthographic branch, and a discourse branch. These
representations are combined through task-specific fusion and
classification components for the primary three-class and secondary
binary tasks.
The semantic target-aware branch represents the complete post, the marked
target span, and target-conditioned surrounding context. The orthographic
branch complements these contextual representations with character-level
and handcrafted writing-style information. The discourse branch provides
global and target-local discourse cues whose contribution is controlled
through a learned gate. In the complete Gated Discourse-OrthoTarget
configuration, these three sources of information are fused into
$z_{\mathrm{Gated}}$. For the three-class task, the fused representation is
passed to a three-unit classifier. For the binary task, the same gated
representation is passed to a two-unit classifier, and the predicted
probability of the \textit{Stigmatizing} class is converted to a final label
using a validation-selected decision threshold. Both the three-class and binary systems are trained with cross-entropy.

The following subsections describe the components of StutterStigma-BERT,
including the semantic target-aware branch, orthographic branch, discourse
branch, task-specific feature fusion and classifiers, and training objectives.

\subsubsection{Semantic Target-Aware Branch}

The semantic target-aware branch represents the complete post, the marked stuttering-related target expression, and the surrounding non-target context within a shared BERTurk encoding. It receives the target-marked sequence defined in Eq.~(\ref{eq:target_marked_input}) and produces three complementary contextual representations: the global post representation $c$, the target-span representation $t$, and the target-conditioned attended context representation $a$. These representations are subsequently combined with interaction features to form the semantic representation used by the downstream branches of StutterStigma-BERT.

\begin{figure*}[t]
    \centering
    \includegraphics[width=\textwidth]{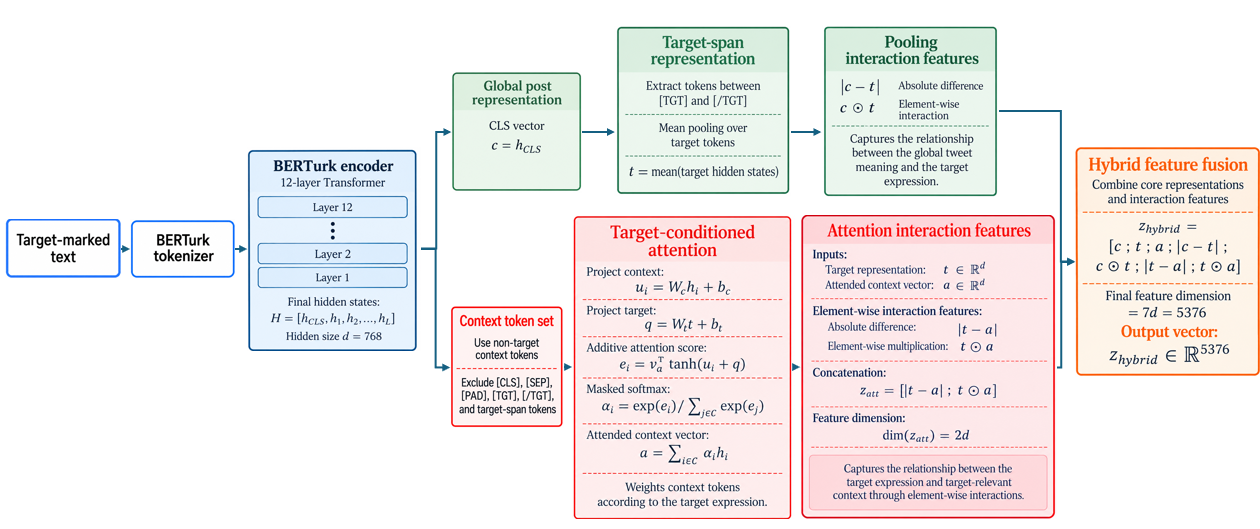}
    \caption{Semantic target-aware branch of StutterStigma-BERT, which extracts pooling interaction features and attention interaction features and combines them with the complete post $c$, the pooled contextual representation of the retained marked target token positions $t$, and the target-conditioned surrounding context $a$ to construct the semantic representation $z_{\mathrm{Hybrid}}$.}
    \label{fig:pool_attention_hybrid}
\end{figure*}

As illustrated in Fig.~\ref{fig:pool_attention_hybrid}, the semantic target-aware
branch first tokenizes the target-marked sequence using the extended BERTurk
tokenizer described in Section~\ref{sec:preprocessing} and then encodes it
with BERTurk. The encoded data from the final-layer hidden states are used in two paths to extract two sets of features: pooling interaction features and attention interaction features. In the first path, to extract the pooling interaction features, the final-layer \texttt{[CLS]} hidden state, $h_{\mathrm{CLS}}$, is
used as the global post representation $c$. For the pooling path, the explicit [TGT] and [/TGT] boundary markers identify the target-token positions used to construct the target-span mask. These positions are then used to obtain the pooled target representation $t$, while the complete
surrounding post context remains available in the BERTurk encoding. The target-span mask identifies all valid token positions located between the corresponding \texttt{[TGT]} and \texttt{[/TGT]} boundary markers, excluding the markers themselves. Let $\mathcal{T}$ denote the set of all retained target-token positions in the input sequence. When a target expression is split into multiple subword units, all of those units are included in $\mathcal{T}$. If a post contains multiple marked target expressions, the valid token positions from all marked spans are pooled jointly rather than first averaging each span separately. The target representation is obtained by mean-pooling the final-layer BERTurk hidden states over all retained target-token positions:

\begin{equation}
t =
\frac{1}{|\mathcal{T}|}
\sum_{i \in \mathcal{T}} h_i ,
\label{eq:target_token_pooling_model}
\end{equation}

where $h_i$ denotes the final-layer BERTurk hidden state at token position $i$, and $|\mathcal{T}|$ denotes the total number of retained target-token positions. Under this formulation, each retained target token contributes equally to the pooled representation. Consequently, when multiple target spans are present, their contribution is proportional to the number of retained subword tokens they contain. In rare cases where no valid target token remains after tokenization or sequence truncation, the mean of the valid non-padding token states is used as a fallback representation. Pooling interaction features are absolute difference and element-wise interaction (Hadamard product) of c and t (i.e., $\left | c-t\right |$ and $c\odot t$, respectively), which capture the relationship between the global post meaning and the target expression.

To extract the attention interaction features in the second path, we first create the valid context token set by excluding the target-span tokens, padding positions, the \texttt{[CLS]} and \texttt{[SEP]} tokens, and the \texttt{[TGT]} and \texttt{[/TGT]} boundary markers:

\begin{equation}
C =
\{i: i \notin \mathcal{T},\ x_i \notin S,\ x_i \neq \mathrm{[PAD]}\},
\label{eq:tca_context_set_model}
\end{equation}

where $S=\{\mathrm{[CLS]},\mathrm{[SEP]},\mathrm{[TGT]},\mathrm{[/TGT]}\}$ denotes the set of special non-context tokens. This surrounding non-target context is represented through target-conditioned attention. The target representation $t$ is used as the conditioning signal, while attention is applied only over valid context tokens. Each valid context-token representation is projected into the attention space as follows:

\begin{equation}
u_i = W_c h_i + b_c,\quad i \in C ,
\label{eq:tca_context_projection_model}
\end{equation}
Here, $h_i$ denotes the final-layer BERTurk hidden state of context token $i$, $W_c$ is the trainable context-projection matrix, and $b_c$ is the corresponding trainable bias vector. The resulting vector $u_i$ is the projected representation of context token $i$ in the attention space. The index $i$ ranges over the valid non-target context positions in $C$. The target representation is then projected into the same
attention space as follows:

\begin{equation}
q = W_t t + b_t .
\label{eq:tca_target_projection_model}
\end{equation}
Here, $t$ is the pooled target-span representation defined in Eq.~(\ref{eq:target_token_pooling_model}), $W_t$ is the trainable target-projection matrix, and $b_t$ is its trainable bias vector. The resulting vector $q$ is the projected target representation in the same attention space as $u_i$. An attention score is then computed for each valid context token as follows:

\begin{equation}
e_i =
v_a^{\top}
\tanh(u_i + q),
\quad i \in C .
\label{eq:tca_attention_score_model}
\end{equation}
Here, $v_a$ is a trainable attention vector, $\tanh(\cdot)$ denotes the hyperbolic tangent activation function, and $e_i$ is the scalar attention score assigned to context token $i$. The score is computed from the interaction between the projected context representation $u_i$ and the projected target representation $q$. The scores are normalized over the valid context set as follows:

\begin{equation}
\alpha_i =
\frac{\exp(e_i)}
{\sum_{j \in C} \exp(e_j)},
\quad i \in C ,
\label{eq:tca_attention_weights_model}
\end{equation}

\begin{figure*}[t]
    \centering
    \includegraphics[width=\textwidth]{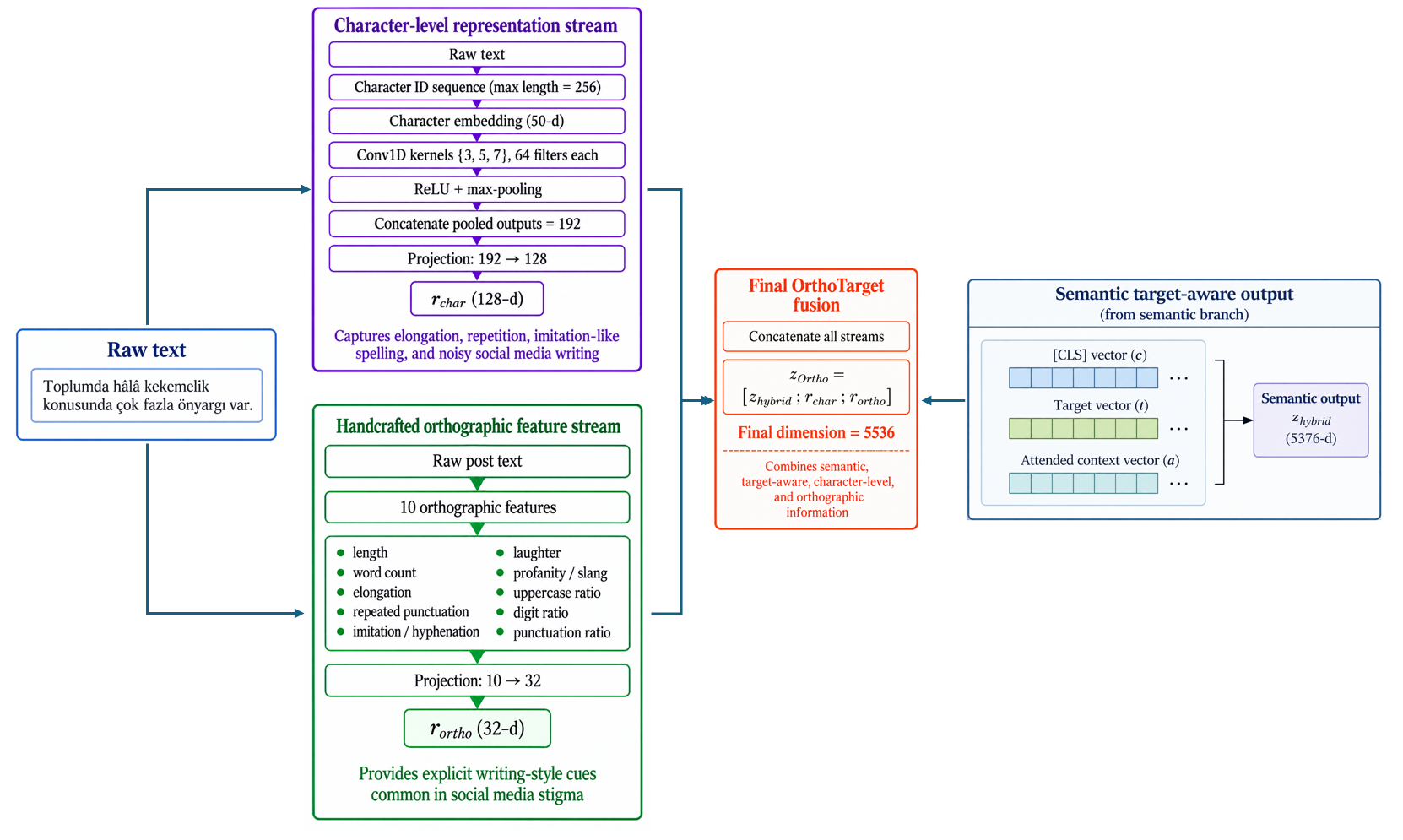}
 \caption{Architecture of the Orthographic branch in our proposed
StutterStigma-BERT framework. The raw post text is processed independently by the
character-level and handcrafted orthographic streams to obtain
$r_{\mathrm{char}}$ and $r_{\mathrm{ortho}}$, respectively. $z_{\mathrm{Hybrid}}$, produced by the semantic target-aware branch (Fig.~\ref{fig:pool_attention_hybrid}), is then combined with these two
representations in the final OrthoTarget fusion component to obtain
$z_{\mathrm{Ortho}}$.}

    \label{fig:orthotarget_architecture}
\end{figure*}

Here, $\alpha_i$ denotes the normalized attention weight assigned to context token $i$, and $\exp(\cdot)$ denotes the exponential function. The denominator sums the attention scores over all valid context positions $j$ in $C$, so the attention weights over the context set sum to one. The attended context representation is then obtained as:

\begin{equation}
a =
\sum_{i \in C}
\alpha_i h_i .
\label{eq:tca_attended_context_model}
\end{equation}
Here, $a$ denotes the target-conditioned attended context representation. It is computed as the weighted sum of the original final layer context-token hidden states $h_i$, where each hidden state is weighted by its normalized attention weight $\alpha_i$. The resulting vector $a$ represents the surrounding non-target context according to its relevance to the marked stuttering-related target. Attention interaction features are absolute difference and element-wise interaction (Hadamard product) of t and a (i.e., $\left | t-a\right |$ and $t\odot a$, respectively), which capture the relationship between the target expression and target-relevant context.

In the semantic target-aware branch of our proposed StutterStigma-BERT framework, $c$ represents the complete post, $t$ represents the pooled contextual representation of the retained marked target token positions, and $a$ represents the target-conditioned surrounding context. These three contextual representations are combined with their pairwise interactions as pooling interaction features and attention interaction features to form the semantic target-aware representation of $z_{\mathrm{Hybrid}}$ as follows:

\begin{equation}
z_{\mathrm{Hybrid}} =
[c;\ t;\ a;\ |c-t|;\ c \odot t;\ |t-a|;\ t \odot a].
\label{eq:hybrid_representation}
\end{equation}

Because each BERTurk-based component is 768-dimensional, $z_{\mathrm{Hybrid}}$ has 5376 dimensions. The output of semantic target-aware branch is the semantic representation $z_{\mathrm{Hybrid}}$, which is subsequently combined with character-level and orthographic representations in the next stage of the framework.


\subsubsection{Orthographic Branch}

The orthographic branch complements the semantic target-aware representation with surface-form information that can capture non-standard and paralinguistic writing patterns in social-media text. As illustrated in Fig.~\ref{fig:orthotarget_architecture}, this branch consists of two parallel components: a character-level representation stream and a handcrafted orthographic feature stream. The outputs of these two streams are subsequently combined with the semantic representation of $z_{\mathrm{Hybrid}}$ from the semantic target-aware branch in the final OrthoTarget fusion component to create the OrthoTarget output vector of $z_{\mathrm{Ortho}}$. These two streams are explained in detail as follows.

\begin{table*}[t]
\caption{Handcrafted Orthographic Features Used in the Orthographic Branch}
\label{tab:orthographic_features}
\centering
\footnotesize
\setlength{\tabcolsep}{3pt}
\renewcommand{\arraystretch}{1.08}
\begin{tabular}{@{}p{0.24\textwidth}p{0.68\textwidth}@{}}
\hline
\textbf{Feature} & \textbf{Definition} \\
\hline
Character length & Post length capped at 300 characters and normalized by 300. \\
Word count & Number of words capped at 60 and normalized by 60. \\
Elongated character count & Count of repeated alphabetic sequences with three or more repetitions, capped and normalized. \\
Repeated punctuation count & Count of repeated punctuation sequences such as ``!!!'', ``???'', or ``...''. \\
Hyphen/mockery pattern & Binary indicator for imitation-like forms such as repeated or hyphenated syllable patterns. \\
Laughter indicator & Binary indicator for laughter forms such as ``ahaha'', ``hahaha'', ``sjsj'', ``jsjs'', ``xd'', or laughing emoji. \\
Profanity indicator & Binary indicator for selected slang or profanity forms used as aggressive or insulting cues. \\
Uppercase ratio & Ratio of uppercase alphabetic characters to all alphabetic characters. \\
Digit ratio & Digit count relative to post length, scaled and capped. \\
Punctuation ratio & Punctuation count relative to post length, scaled and capped. \\
\hline
\end{tabular}
\end{table*}

\textbullet~\textit{\textbf{Character-level representation stream.}}
The character-level stream uses the raw post text as its input so that character-level surface cues, including capitalization, repeated letters, punctuation, and imitation-like spellings, are preserved. Then each post is converted into a fixed-length character sequence with a maximum length of 256. Characters are mapped to trainable 50-dimensional embeddings. The embedded sequence is processed by three one-dimensional convolutional branches with kernel sizes of 3, 5, and 7, each containing 64 filters. Global max pooling is applied to the output of each convolutional branch, and the resulting vectors are concatenated to form a 192-dimensional representation. This representation is then projected into a 128-dimensional character-level vector of $r_{\mathrm{char}}$. Character-level convolutional representations are suitable for capturing local spelling patterns and non-standard surface variation that may not be represented reliably at the token level \cite{zhang_etal_2015_charcnn}. In our proposed StutterStigma-BERT framework, this stream preserves information such as elongated forms, fragmented or imitation-like spellings, and other character-level patterns associated with informal social-media writing.

\textbullet~\textit{\textbf{Handcrafted orthographic feature stream.}}
The second component represents ten explicit writing-style features extracted from the post text, which are all defined in Table~\ref{tab:orthographic_features}. These features capture surface cues that may accompany stuttering-related discourse, including character length, word count, elongated character count, repeated punctuation count, hyphen/mockery pattern, laughter indicator, profanity indicator, uppercase ratio, digit ratio, and punctuation ratio. The resulting ten-dimensional handcrafted feature vector is projected into a 32-dimensional neural representation $r_{\mathrm{ortho}}$. The character-level representation of $r_{\mathrm{char}}$ from the first stream and the handcrafted orthographic representation of $r_{\mathrm{ortho}}$ from the second stream therefore provide complementary learned and explicitly defined surface-form information.

\begin{figure*}[t]
    \centering
    \includegraphics[width=0.85\textwidth]{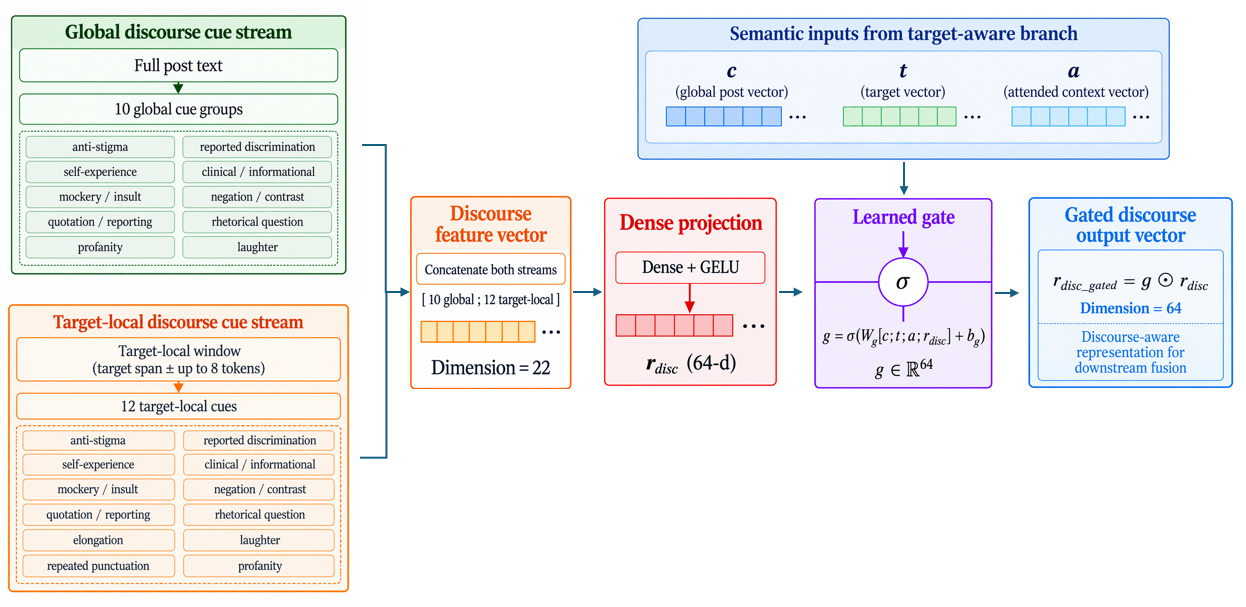}
    \caption{Discourse branch of StutterStigma-BERT, combining global and target-local discourse cues with semantic inputs through a learned gate to produce the gated discourse representation.}
    \label{fig:discourse_branch}
\end{figure*}

\textbullet~\textit{\textbf{OrthoTarget Output Vector.}}
The outputs of the orthographic branch, $r_{\mathrm{char}}$ and $r_{\mathrm{ortho}}$, are combined with the semantic target-aware representation $z_{\mathrm{Hybrid}}$ to construct the OrthoTarget output vector as:

\begin{equation}
z_{\mathrm{Ortho}} =
[z_{\mathrm{Hybrid}};\ r_{\mathrm{char}};\ r_{\mathrm{ortho}}].
\label{eq:orthotarget_fusion_model}
\end{equation}

Because $z_{\mathrm{Hybrid}}$ has 5376 dimensions, $r_{\mathrm{char}}$ has 128 dimensions, and $r_{\mathrm{ortho}}$ has 32 dimensions, the resulting OrthoTarget representation $z_{\mathrm{Ortho}}$ has 5536 dimensions. The character-level stream is learned jointly with the other trainable components of the framework. The ten handcrafted orthographic feature values themselves are deterministically extracted from the post text and are not learned. However, the $10\rightarrow32$ projection that transforms these features into $r_{\mathrm{ortho}}$ is trainable. During model
training, the character-level parameters, the orthographic projection, the trainable BERTurk components, and the downstream classifier are optimized jointly through the task loss. 
$z_{\mathrm{Ortho}}$ is subsequently used by the task-specific components described in the following subsections.

\subsubsection{Discourse Branch}

The discourse branch provides complementary global and target-local
discourse information for the Gated Discourse-OrthoTarget configurations.
It represents discourse-level cues that may help distinguish the role of
stuttering-related language across both the three-class and binary tasks.
These cues are treated as auxiliary input signals rather than class labels or
deterministic decision rules.

As illustrated in Fig.~\ref{fig:discourse_branch}, the discourse representation comprises 22 binary indicators derived from two streams:
a global discourse-cue stream that extracts 10 cues from the complete post,
and a target-local discourse-cue stream that extracts 12 cues from the text
surrounding the marked target expression.
The ten cue groups shared across the global and target-local streams are
anti-stigma language, reported discrimination, self-experience, clinical or informational language, mockery or insult, negation or contrast, reporting or quotation, rhetorical questioning, laughter, and profanity. Each of these ten groups is computed both over the complete post in global discourse cue stream and within the target-local text in target-local discourse cue stream, producing in total 20 indicators. Two additional indicators, elongation and repeated punctuation, are computed only for the target-local representation. 

For each marked stuttering-related target, the target-local window contains the complete marked target span together with up to eight whitespace-delimited tokens immediately preceding it
and up to eight tokens immediately following it. If a post contains more than one marked target span, a local window is constructed for each target and the resulting windows are combined before the target-local discourse cues are extracted. Each discourse cue is represented as a binary indicator: it is set to 1 when at least one predefined pattern belonging to that cue is detected in the relevant text and to 0 otherwise. For the global stream, the relevant text is the complete post, and for the target-local stream, it is the combined target-local window or windows.
The resulting 22-dimensional discourse feature vector from both streams is projected into a 64-dimensional neural representation $r_{\mathrm{disc}}$. A learned gate then controls the contribution of this discourse representation according to the semantic representations of the post. The gate is computed from the global post representation $c$, target-span representation $t$, target-conditioned attended context representation $a$ extracted from semantic target-aware branch, and projected discourse representation of $r_{\mathrm{disc}}$:

\begin{equation}
g =
\sigma
\left(
W_g[c;\ t;\ a;\ r_{\mathrm{disc}}] + b_g
\right),
\quad
g \in \mathbb{R}^{64}
\label{eq:discourse_gate_model}
\end{equation}
Here, $W_g$ denotes the trainable gating weight matrix and $b_g$ denotes the trainable gating bias vector. The concatenated input $[c;t;a;r_{\mathrm{disc}}]$ contains the 768-dimensional global post representation $c$, the 768-dimensional target representation $t$, the 768-dimensional attended-context representation $a$, and the 64-dimensional projected discourse representation $r_{\mathrm{disc}}$. Therefore, $W_g\in\mathbb{R}^{64\times2368}$ and $b_g\in\mathbb{R}^{64}$. The sigmoid function $\sigma(\cdot)$ is applied element-wise to produce the 64-dimensional gating vector $g$. The gated discourse representation is obtained through element-wise multiplication of $g$ and $r_{\mathrm{disc}}$:

\begin{equation}
r_{\mathrm{disc\_gated}}
=
g \odot r_{\mathrm{disc}},
\quad
r_{\mathrm{disc\_gated}} \in \mathbb{R}^{64}
\label{eq:gated_discourse_representation_model}
\end{equation}

The learned gate allows the contribution of the discourse cues to vary
across posts. This is useful because the same surface cue can serve different
discourse functions. Negative language, for example, may occur in a
stigmatizing post but may also appear in supportive counter-speech, while
quotation may indicate reported stigma rather than endorsement. The output
of this branch is the 64-dimensional gated discourse representation
$r_{\mathrm{disc\_gated}}$, which is combined with the semantic and
orthographic representations to form the final StutterStigma-BERT configuration used in the final three-class
and binary systems.

\subsubsection{Feature Fusion and Classifier for Three-Class Classification}

For three-class classification, feature fusion extends the semantic representation defined in the semantic target-aware branch by additionally incorporating interactions between the global post representation $c$ and the target-conditioned attended context representation $a$. The expanded semantic representation is defined as:

\begin{equation}
z_{\mathrm{sem\_v2}} =
[c;\ t;\ a;\ |c-t|;\ c \odot t;\ |t-a|;\ t \odot a;\ |c-a|;\ c \odot a].
\label{eq:expanded_semantic_representation_model}
\end{equation}

Because each BERTurk-based vector is 768-dimensional, the expanded semantic representation $z_{\mathrm{sem\_v2}}$ has 6912 dimensions. Relative to $z_{\mathrm{Hybrid}}$, this representation retains the same global, target-span, and target-conditioned context vectors while additionally modelling the interaction between the complete post representation and the attended contextual representation through $|c-a|$ and $c \odot a$. The expanded semantic representation is then combined with the 128-dimensional character-level representation $r_{\mathrm{char}}$, the 32-dimensional handcrafted orthographic representation $r_{\mathrm{ortho}}$, and the 64-dimensional gated discourse representation $r_{\mathrm{disc\_gated}}$ as follows:

\begin{equation}
z_{\mathrm{Gated}} =
[z_{\mathrm{sem\_v2}};\ r_{\mathrm{char}};\ r_{\mathrm{ortho}};\ r_{\mathrm{disc\_gated}}].
\label{eq:gated_discourse_fusion_model}
\end{equation}

The resulting fused representation $z_{\mathrm{Gated}}$ has 7136 dimensions and constitutes the input to the three-class classifier. Our three-class classifier consists of a dropout layer followed by a fully connected layer with GELU activation, a second dropout layer, and a final linear output layer with three units. These three output units correspond to the \textit{Stigmatizing}, \textit{Non-stigmatizing/contextual}, and \textit{Supportive} classes. 



\subsubsection{Loss Function for Three-Class Classification}

Cross-entropy (CE) is used as the loss function for the
three-class StutterStigma-BERT model. For a training example with target
distribution $\tilde{y}$ and predicted class probabilities $p$, the
cross-entropy loss is defined as:

\begin{equation}
\mathcal{L}_{\mathrm{CE}}
=
-\sum_{k=1}^{K}
\tilde{y}_{k}\log p_{k}
\label{eq:cross_entropy_loss}
\end{equation}

where $K$ denotes the number of classes, $\tilde{y}_{k}$ denotes the target
value for class $k$, and $p_{k}$ denotes the predicted probability assigned
to that class. For the three-class task, $K=3$. Under the common training
protocol, label smoothing with $\epsilon=0.05$ is applied to the target
distribution. The classifier parameters and all trainable upstream components
are optimized jointly through this classification objective. R-Drop consistency
regularization is applied separately as described in
Section~\ref{sec:experiment_setup}.

\subsubsection{Feature Fusion, Classifier, and Loss Function for Binary Classification}

For the secondary binary task, the original
\textit{Non-stigmatizing/ contextual} and \textit{Supportive} classes are
merged into a single \textit{Non-stigma} category, while
\textit{Stigmatizing} remains the positive class. The final binary
configuration uses $z_{\mathrm{Gated}}$ defined in
Eq.~(\ref{eq:gated_discourse_fusion_model}) for classification. This representation combines
the expanded semantic target-aware features, character-level
representation, handcrafted orthographic representation, and gated
discourse representation.

The fused representation of $z_{\mathrm{Gated}}$ is passed to a binary classifier consisting of a
dropout layer, a fully connected layer with GELU activation, a second
dropout layer, and a final linear output layer with two units corresponding
to \textit{Stigmatizing} and \textit{Non-stigma}. The binary classifier is
trained with the same cross-entropy objective defined in
Eq.~(\ref{eq:cross_entropy_loss}), with $K=2$ for the
\textit{Stigmatizing} and \textit{Non-stigma} classes.
During inference, the predicted probability of the positive
\textit{Stigmatizing} class for post $i$, denoted by $p_i$, is converted
into a binary label using the validation-selected decision threshold
$\theta^{*}$:

\begin{equation}
\hat{y}_i =
\begin{cases}
1, & p_i \geq \theta^{*},\\
0, & p_i < \theta^{*},
\end{cases}
\label{eq:binary_decision_rule}
\end{equation}

where $\hat{y}_i=1$ denotes \textit{Stigmatizing} and
$\hat{y}_i=0$ denotes \textit{Non-stigma}. The threshold $\theta^{*}$ is
selected independently from the corresponding validation split and is
subsequently applied unchanged to the held-out test probabilities. The candidate threshold range and optimization rule are specified in
Section~\ref{sec:binary_threshold_sensitivity}.

\section{Experiments and Results}
\label{sec:experiments_results}

This section describes the experimental configuration and evaluates
StutterStigma-BERT across complementary three-class and binary settings.
We first summarize the experimental setup, evaluation metrics, and
training diagnostics, and then report the final 20-seed performance, the
component and backbone ablations, the robustness analysis, threshold
sensitivity, baseline comparisons, zero-shot LLM evaluation, and error
analysis.

\subsection{Experiment Setup and Hyperparameters}
\label{sec:experiment_setup}

All in-domain neural models use the same fixed model-ready dataset and
held-out test partition. The training partition contains 4,815 posts, and the
held-out test partition contains 1,204 posts. The StutterStigma-BERT
configurations, alternative-backbone analyses, and task-adapted neural
comparators are trained using the common 20-epoch protocol described below,
with seed-specific stratified 90/10 internal-training and validation splits
and validation-based checkpoint selection, with experiment-specific adaptations described in their corresponding
analysis subsections.
The TF--IDF baselines are fitted on the same fixed training partition using
their model-appropriate optimization procedures. The direct-transfer
harmful-language classifiers and zero-shot LLMs are evaluated on the same
fixed held-out test set without training on StutterStigma-TR. 

Within each random-seed run, the training partition is further divided into a seed-specific stratified internal-training subset and validation subset corresponding to 90\% and 10\% of the training partition, respectively. This produces 4,333 internal-training posts and 482 validation posts per run. The held-out test partition remains unchanged across all model families and is not used for gradient fitting, checkpoint selection, regularization selection, or decision-threshold selection. The neural experiments are conducted on an NVIDIA A100-SXM4-40GB GPU
under Linux, using Python 3.13.15 and PyTorch 2.11.0 with CUDA 12.8.

\begin{table}[t]
\centering
\caption{
Final Training Configuration and Hyperparameters for the Selected StutterStigma-BERT Systems (binary and three-class classification schemes)}
\label{tab:final_training_configurations}

{
\small
\setlength{\tabcolsep}{6pt}
\renewcommand{\arraystretch}{1.08}

\begin{tabular}{@{}lc@{}}
\toprule
\textbf{Component} &
\textbf{Configuration} \\
\midrule


Training objective &
Cross-entropy \\

Maximum epochs &
20 \\

Checkpoint selection &
Best validation Macro-F1 \\

Optimizer &
AdamW \\

Learning rate &
$2\times10^{-5}$ \\

LR schedule &
Linear \\

Warm-up ratio &
0.20 \\

Task dropout &
0.20 \\

Encoder dropout &
0.15 \\

Weight decay &
0.05 \\

Label smoothing &
0.05 \\

R-Drop &
$\alpha_{\mathrm{RDrop}}=1.0$ \\

Frozen encoder blocks &
Bottom 4 of 12 \\

Training / evaluation batch size &
16 / 32 \\

Maximum token length &
160 \\

Maximum character length &
256 \\

Ablation seeds &
42, 123, 2026 \\

Final robustness evaluation &
20 distinct seeds \\

\bottomrule
\end{tabular}

\vspace{1mm}
\begin{minipage}{0.95\columnwidth}
\footnotesize
\textit{Note.}
The table summarizes the common training configuration used for the
validation-selected three-class and binary StutterStigma-BERT systems.
The controlled ablations use the same training and model-selection
protocol, with architectural components varying by configuration.
\end{minipage}
}

\end{table}

Raw BERTurk and the StutterStigma-BERT configurations use
\texttt{dbmdz/bert-base-turkish-cased}, the cased BERTurk encoder
\cite{schweter_2020_berturk}. For the controlled BERTurk-based
configurations, the tokenizer vocabulary includes the four special tokens
\texttt{[TGT]}, \texttt{[/TGT]}, \texttt{[USER]}, and \texttt{[URL]}, and
the embedding matrix is resized accordingly before training. Raw BERTurk
receives \texttt{preprocessed\_text}, which contains no target-boundary
markers, whereas the target-aware configurations receive
\texttt{target\_marked\_text}. 

All BERTurk-based neural models are initialized from the pretrained
\texttt{dbmdz/bert-base-turkish-cased} checkpoint \cite{schweter_2020_berturk} rather than trained
from scratch. During fine-tuning, the bottom four of the 12 Transformer
encoder blocks (layers 0--3 in zero-based indexing) are frozen, while the
embedding layer and the upper eight encoder blocks (layers 4--11) remain
trainable. The resized embedding matrix, including the newly added
\texttt{[TGT]}, \texttt{[/TGT]}, \texttt{[USER]}, and \texttt{[URL]}
token embeddings, is optimized jointly with the trainable encoder blocks
and the task-specific components. In the final three-class StutterStigma-BERT framework, these components include the
target-conditioned attention layers, character-level CNN, orthographic
projection, discourse projection and gating network, and the final
classifier. This fine-tuning strategy preserves the pretrained lower
encoder blocks while allowing the remaining encoder and task-specific
parameters to adapt to StutterStigma-TR dataset.

R-Drop regularization is implemented by passing each training example
through the model twice under independently sampled dropout masks and
minimizing the symmetric Kullback--Leibler divergence between the two
resulting output distributions, in addition to the classification loss
\cite{wu2021r}. In the common training protocol, R-Drop is applied
with $\alpha_{\mathrm{RDrop}}=1.0$. Cross-entropy is used together with
label smoothing ($\epsilon=0.05$) throughout the common neural training
protocol. 

The controlled ablation analyses are conducted across three random seeds
(42, 123, and 2026). Within each run, validation is performed after every
epoch, and the checkpoint with the highest internal-validation Macro-F1 is
retained. Model configuration is determined from validation performance,
whereas the fixed held-out test set is used only for final evaluation of the
corresponding validation-selected checkpoint.
The controlled neural analyses use a common 20-epoch training and
model-selection protocol wherever the corresponding settings are applicable.
The common protocol uses the same fixed data split, random seeds, optimizer,
learning rate, warm-up ratio, weight decay, task dropout, label smoothing,
R-Drop coefficient, batch sizes, and validation-based checkpoint-selection
criterion. Experiment-specific architectural adaptations are described in
their corresponding analysis subsections.

Table~\ref{tab:final_training_configurations} summarizes the final training
configurations and final hyperparameters selected for the three-class and binary StutterStigma-BERT
systems.
Several controls are applied to reduce information leakage. Duplicate removal is completed before the train--test split. The same fixed held-out test set is reused across model families so that within-study comparisons are based on identical test examples. The character vocabulary is constructed only from the corresponding training data within each random-seed run. Model configuration, regularization, checkpoint, and binary threshold decisions are based on internal-validation data rather than held-out test labels.

\subsection{Evaluation Metrics}

Model performance is evaluated separately for the primary three-class classification and the secondary binary classification tasks. The primary evaluation metric for both tasks is Macro-F1. Macro-F1 is selected because the dataset is not fully balanced and because each discourse category is theoretically important. 
Macro-F1 is used as the primary class-balanced performance measure,
with Accuracy and Weighted-F1 reported as complementary aggregate metrics.
For each class $c$, precision and recall are defined as:

\begin{equation}
Precision_{c} =
\frac{{TP}_{c}}{{TP}_{c}+{FP}_{c}}
\label{eq:precision}
\end{equation}

\begin{equation}
{Recall}_{c} =
\frac{{TP}_{c}}{{TP}_{c}+{FN}_{c}}
\label{eq:recall}
\end{equation}

Here, $TP_c$, $FP_c$, $FN_c$, and $TN_c$ denote the numbers of
true-positive, false-positive, false-negative, and true-negative predictions
for class $c$, respectively. These quantities are defined in a one-vs.-rest
manner for each class.
Macro-Precision and Macro-Recall are obtained by taking the unweighted
mean of the corresponding class-specific precision and recall values:

\begin{equation}
{MacroPrecision}
=
\frac{1}{|C|}
\sum_{c \in C} {Precision}_{c}
\label{eq:macro_precision}
\end{equation}

\begin{equation}
{MacroRecall}
=
\frac{1}{|C|}
\sum_{c \in C} {Recall}_{c}
\label{eq:macro_recall}
\end{equation}

\begin{figure*}[!t]
\centering

\begin{minipage}[t]{0.485\textwidth}
\centering
\includegraphics[width=\linewidth]{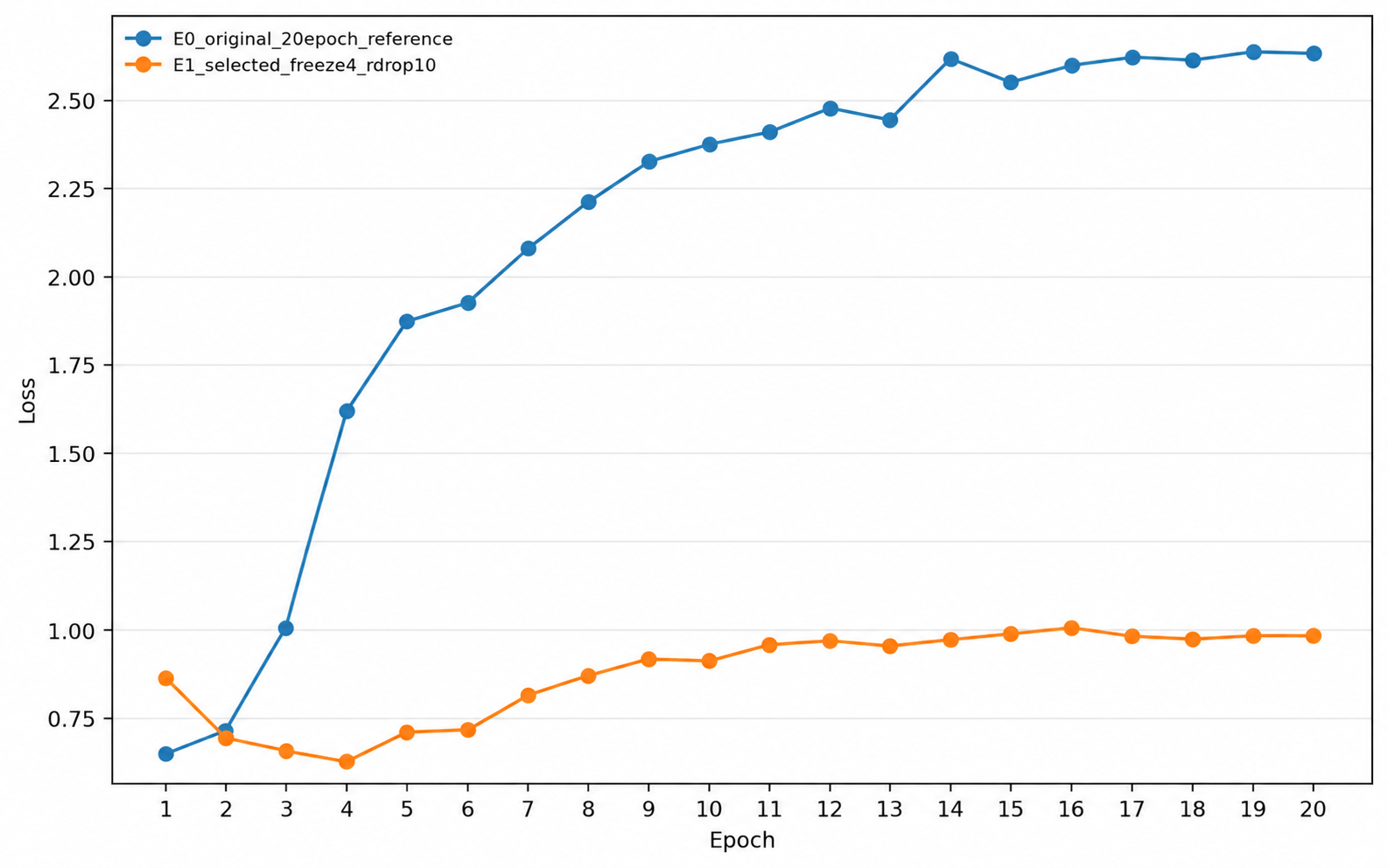}
\vspace{-1mm}

\textbf{(a)} Standard validation NLL
\end{minipage}
\hfill
\begin{minipage}[t]{0.485\textwidth}
\centering
\includegraphics[width=\linewidth]{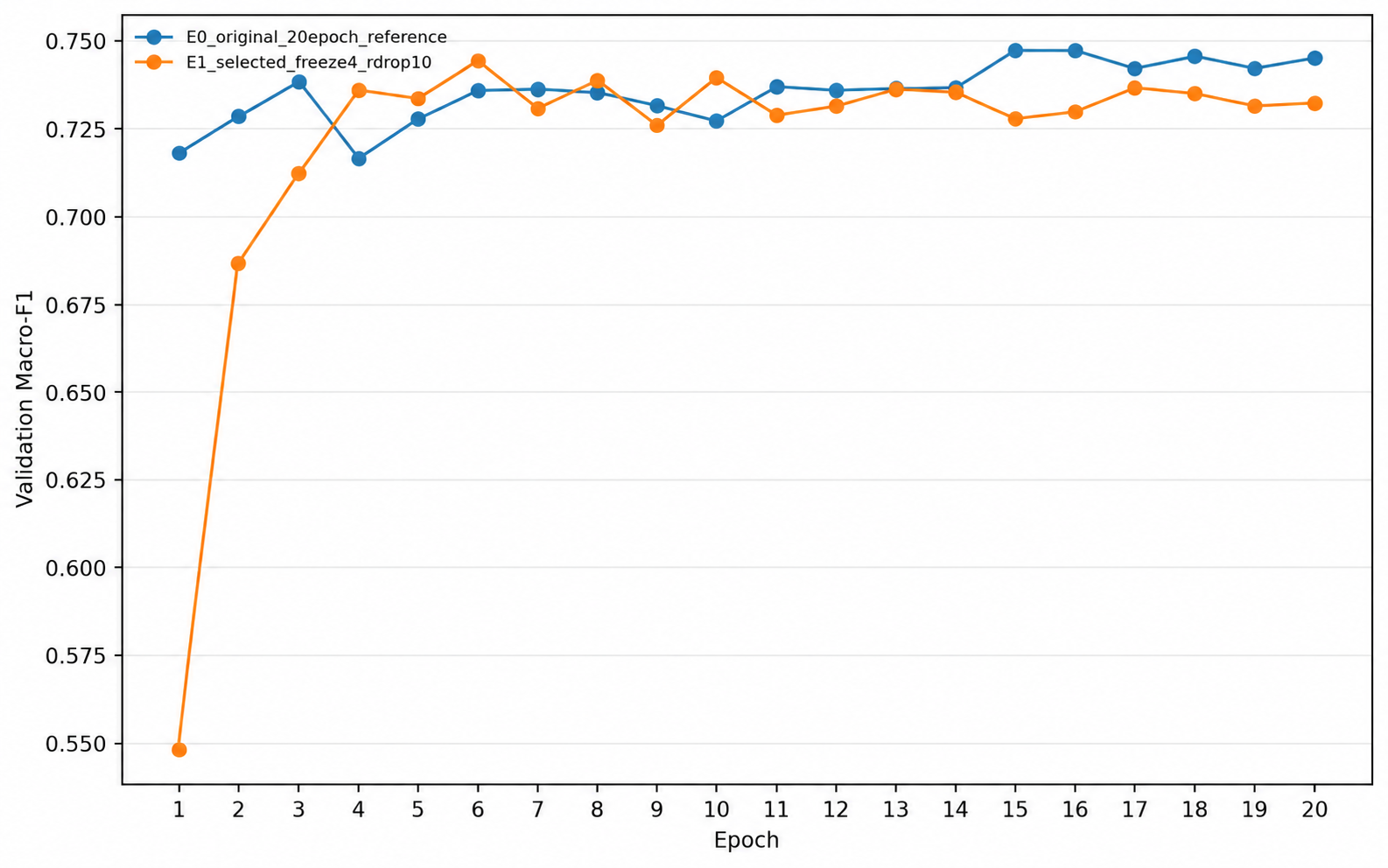}
\vspace{-1mm}

\textbf{(b)} Validation Macro-F1
\end{minipage}

\caption{
Training-dynamics comparison across the 20-epoch schedule under the reference
and selected regularization configurations.
(a) Standard validation negative log-likelihood (NLL).
(b) Validation Macro-F1. Curves summarize the separate five-seed
regularization-confirmation analysis used to compare training stability and
confirm the regularization setting subsequently applied in the common
controlled experiments.
}
\label{fig:regularization_diagnostics}
\end{figure*}

where $C$ denotes the set of class labels. The class-specific F1 score is computed as:
\begin{equation}
{F1}_{c} =
\frac{2 \cdot {Precision}_{c} \cdot {Recall}_{c}}
{{Precision}_{c}+{Recall}_{c}}.
\label{eq:f1}
\end{equation}

Macro-F1 is calculated by averaging the class-specific F1 scores across all classes:
\begin{equation}
{MacroF1} =
\frac{1}{|C|}
\sum_{c \in C} {F1}_{c},
\label{eq:macro_f1}
\end{equation}

where $C$ denotes the set of class labels. For the primary three-class classification task, $|C|=3$ and for the secondary binary classification task, $|C|=2$.
Accuracy is defined as the proportion of correctly classified examples as follows:

\begin{equation}
{Accuracy}
=
\frac{\sum_{c \in C} TP_c}{N},
\label{eq:accuracy}
\end{equation}

where $N$ denotes the total number of evaluated samples.
Weighted-F1 additionally accounts for class frequency by weighting each
class-specific F1 score according to its support:

\begin{equation}
{WeightedF1}
=
\sum_{c \in C}
\frac{n_c}{N}F1_c,
\label{eq:weighted_f1}
\end{equation}

where $n_c$ denotes the number of gold-standard examples belonging to
class $c$. Accuracy and Weighted-F1 are reported as complementary
aggregate measures, whereas Macro-F1 remains the primary class-balanced
evaluation metric.

For the binary classification task, the area under the receiver operating
characteristic curve (ROC-AUC) and the area under the precision--recall
curve (PR-AUC) are additionally reported to characterize
threshold-independent discrimination and the precision--recall trade-off
for the positive \textit{Stigmatizing} class
\cite{sokolova_lapalme_2009,fawcett2006introduction,
davis2006relationship,saito2015precision}.
Threshold-dependent binary metrics are calculated using the
validation-selected decision threshold described in
Section~\ref{sec:binary_threshold_sensitivity}, whereas ROC-AUC and PR-AUC are
computed from the continuous predicted probabilities and therefore do not
depend on the selected classification threshold.

\subsection{Training Diagnostics and Regularization Refinement}
\label{sec:training_diagnostics}

Training dynamics are examined over the full 20-epoch schedule using validation Macro-F1 together with standard validation negative log-likelihood (NLL). 
Standard validation NLL is monitored as a diagnostic of probabilistic fit,
while validation Macro-F1 remains the primary class-balanced criterion for
checkpoint and configuration selection. The two measures are tracked
separately because they characterize different aspects of model behavior
over the training schedule. Validation Negative Log-Likelihood (NLL) is used to assess the quality of the model's probabilistic predictions. For a classification task, NLL is defined as: 

\begin{equation} \mathrm{NLL} = -\frac{1}{N} \sum_{i=1}^{N} \log p(y_i \mid x_i), \end{equation} 

where $N$ denotes the number of validation samples and $p(y_i \mid x_i)$ represents the probability assigned by the model to the true class $y_i$ for input $x_i$.
A five-seed diagnostic analysis compares two 20-epoch regularization
configurations of the Gated Discourse-OrthoTarget model trained with
cross-entropy. The reference and selected configurations are evaluated
using the same five random seeds. Standard validation NLL is used to
characterize probabilistic deterioration across epochs, whereas validation
Macro-F1 is examined separately as the class-balanced predictive criterion.
The analysis evaluates whether the selected regularization strategy delays
and reduces late-epoch validation-loss deterioration without materially
reducing the best observed validation Macro-F1.
Under the reference configuration, the mean epoch of minimum validation NLL
is 1.4, and sustained NLL deterioration of at least 10\% begins at a mean
epoch of 2.8. The mean increase from minimum validation NLL to epoch 20 is
$+2.061$. Under the selected regularization configuration, the mean epoch of
minimum validation NLL shifts to 3.4, sustained deterioration begins at a mean
epoch of 5.8, and the corresponding late-epoch increase decreases to
$+0.301$. Mean best validation Macro-F1 remains closely matched between the
reference and selected configurations, at 75.93\% and 75.76\%, respectively.

Thus, the selected configuration delays and substantially reduces late
validation-loss deterioration without materially changing the best observed
validation Macro-F1. Fig.~\ref{fig:regularization_diagnostics} illustrates
these training dynamics. The validation-NLL curves show a later and markedly
smaller deterioration under the selected regularization strategy, whereas the
validation Macro-F1 curves remain comparatively similar. The refinement
therefore targets training stability rather than an increase in peak
validation classification performance.

\begin{table*}[t]
\caption{Final Model Performance of the Proposed StutterStigma-BERT Model Across 20 Random Seeds on Held-out Test Data for Binary and Three-class Classification}
\label{tab:final_model_performance}
\centering
\scriptsize
\setlength{\tabcolsep}{3.2pt}
\renewcommand{\arraystretch}{1.12}

{
\begin{adjustbox}{max width=\textwidth}
\begin{tabular}{llccccccc}
\toprule
\textbf{Task} &
\textbf{Final configuration} &
\makecell{\textbf{Accuracy}\\\textbf{mean (\%)}} &
\makecell{\textbf{Macro-}\\\textbf{Precision (\%)}} &
\makecell{\textbf{Macro-}\\\textbf{Recall (\%)}} &
\makecell{\textbf{Macro-F1}\\\textbf{mean (\%)}} &
\makecell{\textbf{Weighted-F1}\\\textbf{mean (\%)}} &
\makecell{\textbf{ROC-AUC}\\\textbf{mean (\%)}} &
\makecell{\textbf{PR-AUC}\\\textbf{mean (\%)}} \\
\midrule

Three-class &
\textbf{Our Proposed Three-class StutterStigma-BERT} &
74.36\% &
73.68\% &
73.85\% &
73.54\% &
74.45\% &
-- &
-- \\

Binary &
\textbf{Our Proposed Binary StutterStigma-BERT} &
85.77\% &
85.71\% &
84.84\% &
85.17\% &
85.69\% &
87.83\% &
84.68\% \\

\bottomrule
\end{tabular}
\end{adjustbox}
\vspace{1mm}
\begin{minipage}{0.98\textwidth}
\footnotesize
\textit{Note.} Values are mean percentages across 20 random-seed
evaluations of the validation-selected final configurations on held-out test data.
For the binary task, threshold-dependent metrics are calculated
using the decision threshold selected independently from the corresponding
validation split. ROC-AUC and PR-AUC are computed from continuous predicted
probabilities and are therefore threshold-independent. CE = cross-entropy.
\end{minipage}
}
\end{table*}

\begin{figure*}[t]
\centering

\begin{minipage}[t]{0.49\textwidth}
\centering
\includegraphics[width=\linewidth]{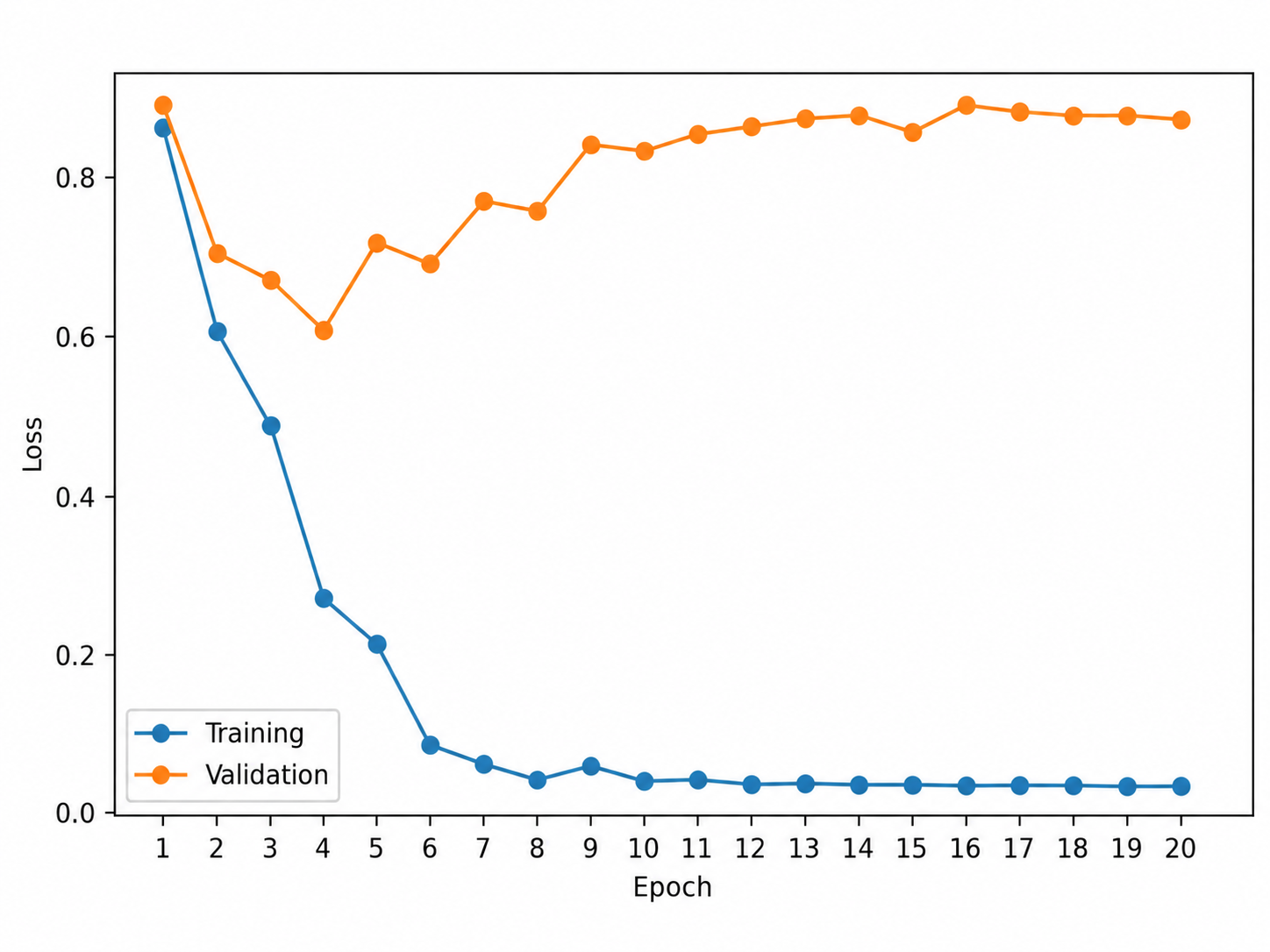}

{\small(a) Training and validation loss vs. epochs}
\end{minipage}
\hfill
\begin{minipage}[t]{0.49\textwidth}
\centering
\includegraphics[width=\linewidth]{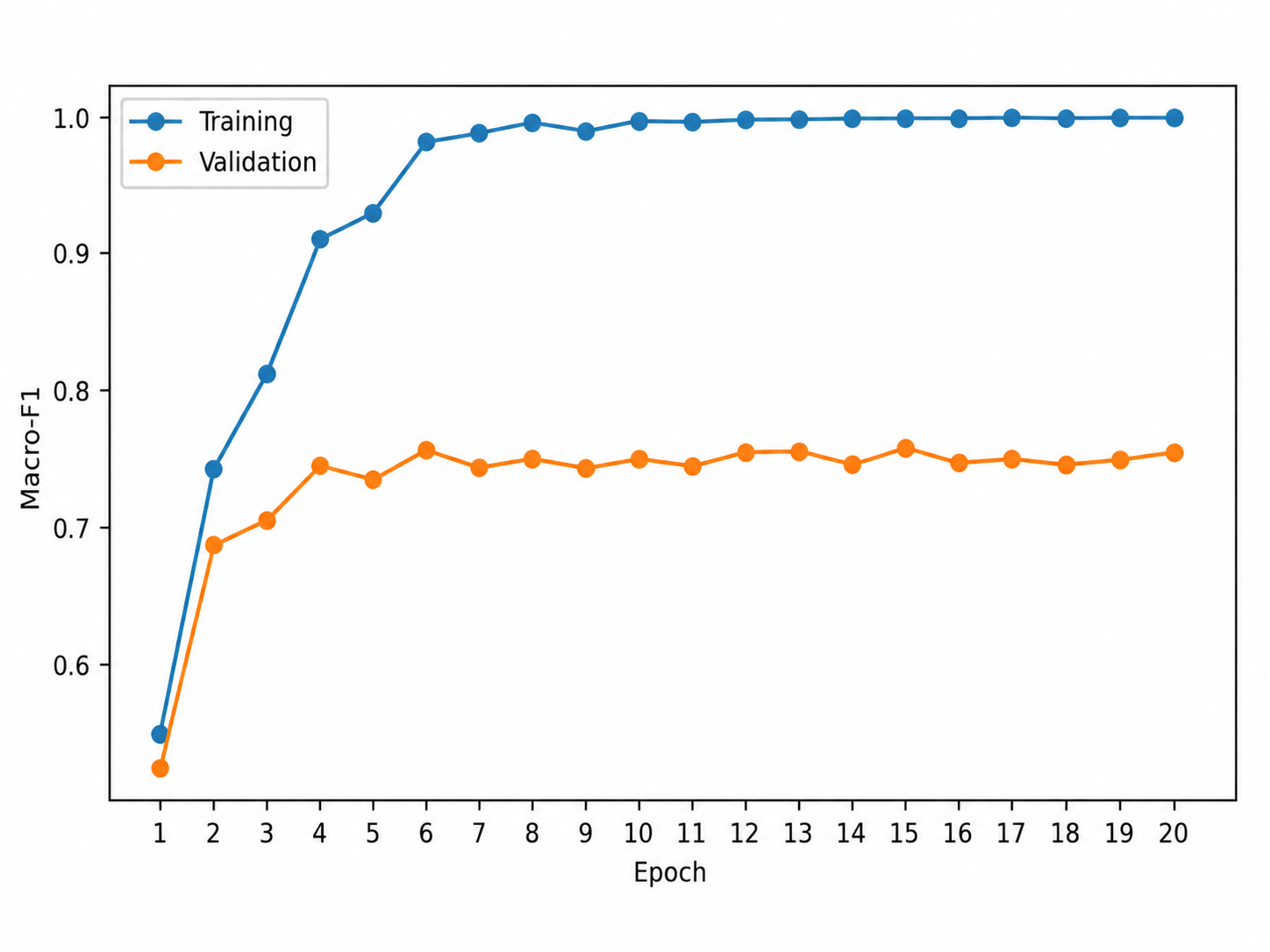}

{\small(b) Training and validation Macro-F1 vs. epochs}
\end{minipage}

\caption{
Learning curves of the final three-class StutterStigma-BERT model across 20 epochs. Curves show epoch-wise means across the three controlled
diagnostic seeds (42, 123, and 2026). (a) Training and validation loss vs. epochs, (b) Training
and validation Macro-F1 vs. epochs.
}
\label{fig:threeclass_final_learning_curves}
\end{figure*}

\begin{figure*}[h!]
\centering

\begin{minipage}[t]{0.49\textwidth}
\centering
\includegraphics[width=\linewidth]{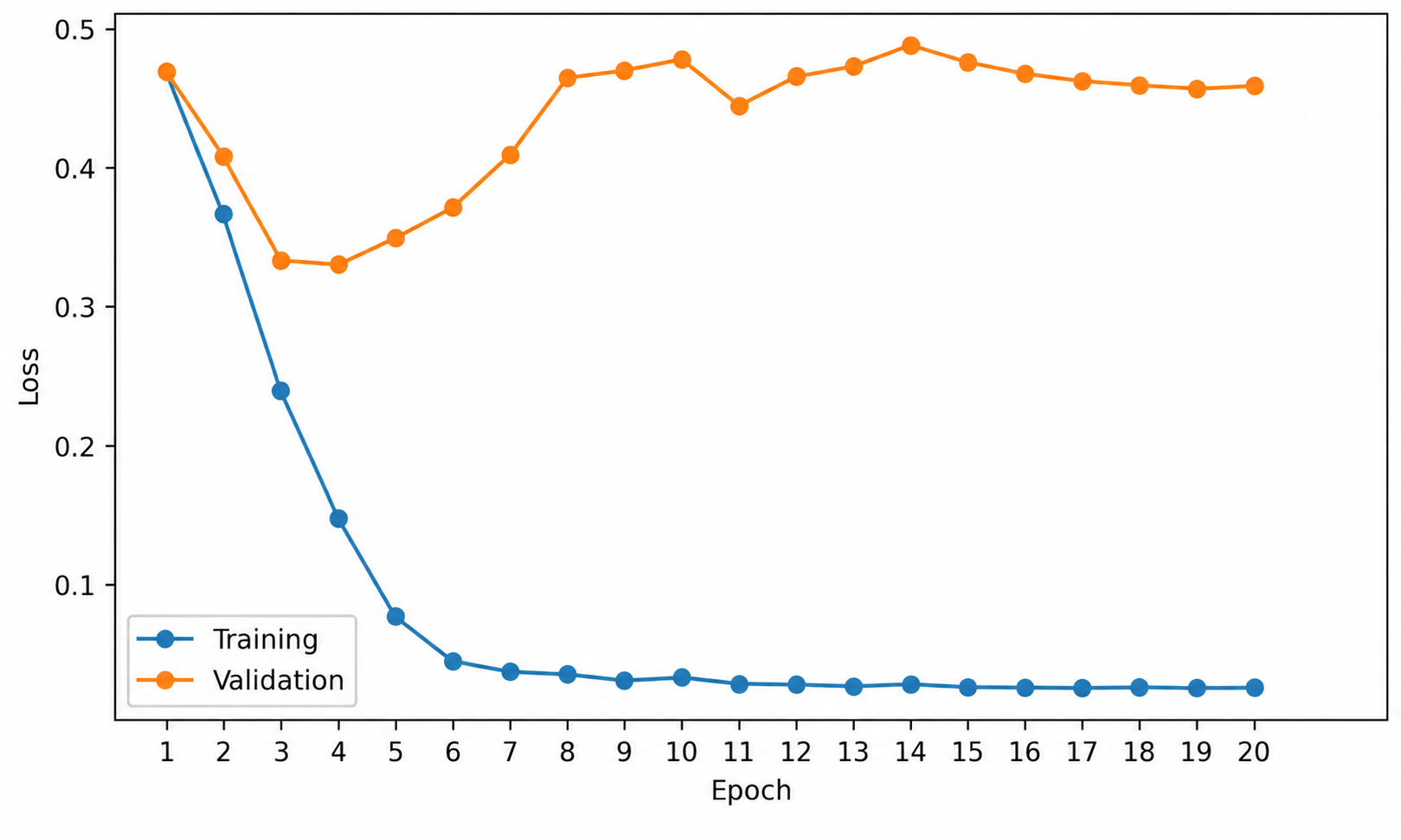}

{\small(a) Training and validation loss vs. epochs}
\end{minipage}
\hfill
\begin{minipage}[t]{0.49\textwidth}
\centering
\includegraphics[width=\linewidth]{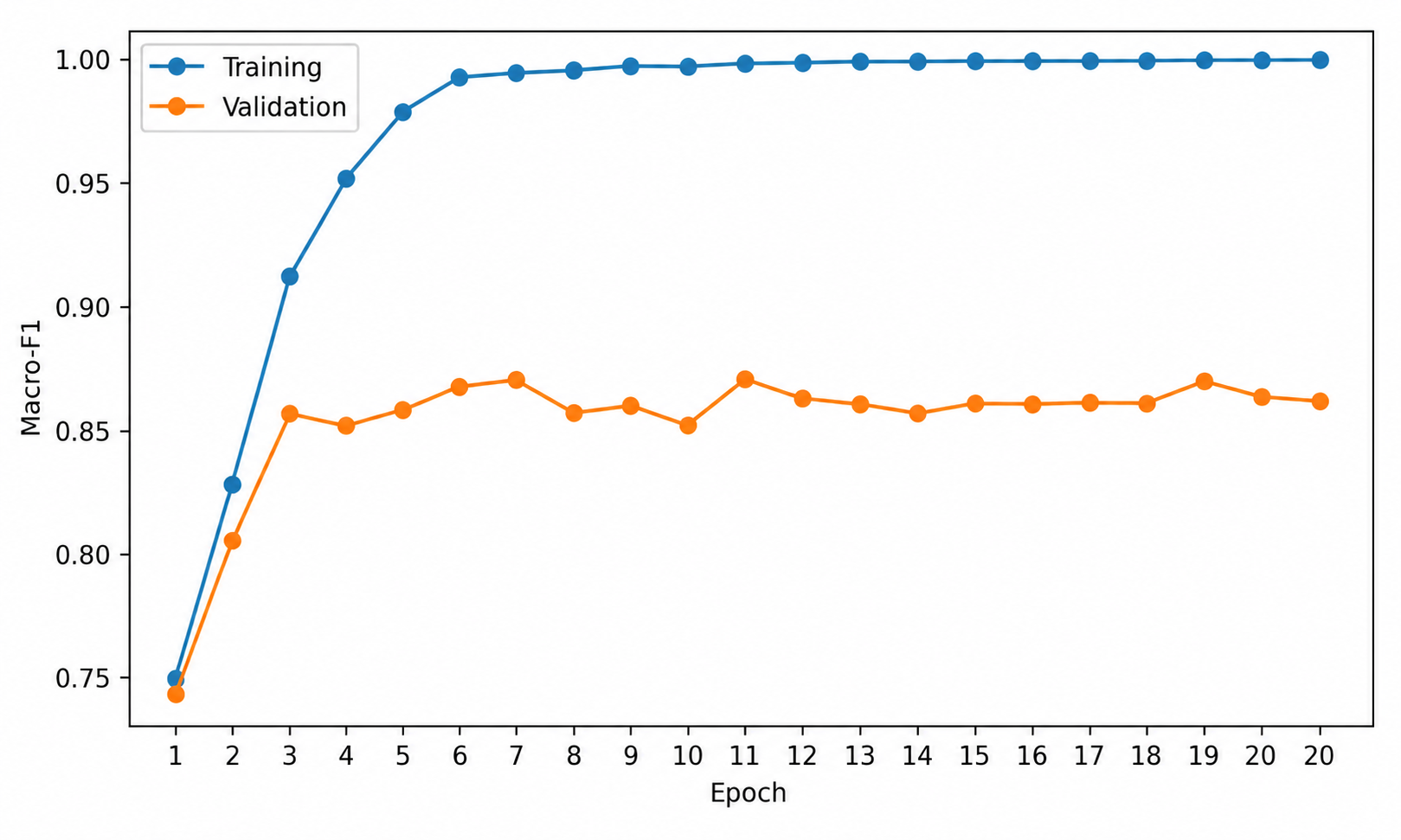}

{\small(b) Training and validation Macro-F1 vs. epochs}
\end{minipage}

\caption{
Learning curves of the final binary StutterStigma-BERT model across 20 epochs. Curves show epoch-wise means across the three controlled
diagnostic seeds (42, 123, and 2026). (a) Training and validation loss vs. epochs, (b) Training
and validation Macro-F1 vs. epochs.
}
\label{fig:binary_final_learning_curves}
\end{figure*}

RAdam \cite{liu2020radam} produces comparatively stable validation-loss
behavior but substantially weaker validation classification performance and is
therefore not retained. Mixout \cite{lee2020mixout} delays the onset of
sustained validation-loss deterioration, but reduces validation Macro-F1 by
approximately one percentage point. The final configuration consequently retains AdamW and does not use Mixout. The selected regularization configuration is retained for the subsequent
controlled ablation analyses and final repeated-seed evaluations. The
five-seed regularization-confirmation analysis is used to compare the
training stability of the reference and selected regularization settings and
to confirm the setting used in the subsequent experiments. Final predictive
performance is evaluated separately using the validation-selected checkpoints
of the final three-class and binary systems.

\begin{table*}[t]
\caption{Controlled Ablation Study on Three-Class
StutterStigma-BERT Model Across Different Model Components Under the Common Training Protocol}
\label{tab:three_class_ablation}
\centering
\scriptsize
\setlength{\tabcolsep}{2.2pt}
\renewcommand{\arraystretch}{1.15}

{
\begin{adjustbox}{max width=\textwidth}
\begin{tabular}{lccccccccccc}
\toprule
\textbf{Model Components} &
\makecell{\textbf{Target}\\\textbf{marking}} &
\makecell{\textbf{Target-span}\\\textbf{pooling}} &
\makecell{\textbf{Target-cond.}\\\textbf{attention}} &
\makecell{\textbf{Char-}\\\textbf{CNN}} &
\makecell{\textbf{Orthographic}\\\textbf{features}} &
\makecell{\textbf{{$c$--$a$}}\\\textbf{interactions}} &
\makecell{\textbf{Gated}\\\textbf{discourse}} &
\makecell{\textbf{Training}\\\textbf{Macro-F1 (\%)}} &
\makecell{\textbf{Training}\\\textbf{Accuracy (\%)}} &
\makecell{\textbf{Validation}\\\textbf{Macro-F1 (\%)}} &
\makecell{\textbf{Validation}\\\textbf{Accuracy (\%)}} \\
\midrule

Raw BERTurk
& -- & -- & -- & -- & -- & -- & --
& 94.40\% & 94.62\% & 75.01\% & 75.73\% \\

BERTurk + \texttt{[TGT]}
& \checkmark & -- & -- & -- & -- & -- & --
& 92.51\% & 92.68\% & 75.37\% & 75.86\% \\

Target-Pooling BERTurk
& \checkmark & \checkmark & -- & -- & -- & -- & --
& \textbf{99.15\%} & \textbf{99.18\%} & 75.81\% & 76.63\% \\

Pool+Attention Hybrid
& \checkmark & \checkmark & \checkmark & -- & -- & -- & --
& 96.87\% & 96.97\% & 75.53\% & 76.49\% \\

OrthoTarget Hybrid CE
& \checkmark & \checkmark & \checkmark & \checkmark & \checkmark & -- & --
& 97.29\% & 97.38\% & 76.05\% & 76.83\% \\

\textbf{StutterStigma-BERT (Ours)}
& \checkmark & \checkmark & \checkmark & \checkmark & \checkmark &
\checkmark & \checkmark
& 99.13\% & \textbf{99.18\%} &
\textbf{76.32\%} & \textbf{77.18\%} \\

\bottomrule
\end{tabular}
\end{adjustbox}

\vspace{1mm}
\begin{minipage}{0.98\textwidth}
\footnotesize
\textit{Note.}
Raw BERTurk uses preprocessed text without target markers and serves as the
contextual reference for the controlled comparison. Values are means across
three random-seed runs (42, 123, and 2026) under the common 20-epoch
training protocol. For each seed, training and validation metrics are measured
at the same checkpoint selected by the highest internal-validation Macro-F1.
Training metrics are reported descriptively, whereas architecture selection is
based on validation Macro-F1. Bold values indicate the highest mean value
within each reported metric (tied maxima are both shown in bold). The
$c$--$a$ interaction column denotes the additional $|c-a|$ and
$c\odot a$ terms in the expanded semantic representation.
CE = cross-entropy.
\end{minipage}
}
\end{table*}

\begin{table*}[t]
\caption{Controlled Ablation Study on Binary
StutterStigma-BERT Model Across Different Model Components Under the Common Training Protocol}
\label{tab:binary_ablation}
\centering
\scriptsize
\setlength{\tabcolsep}{2.2pt}
\renewcommand{\arraystretch}{1.15}

{
\begin{adjustbox}{max width=\textwidth}
\begin{tabular}{lccccccccccc}
\toprule
\textbf{Model Components} &
\makecell{\textbf{Target}\\\textbf{marking}} &
\makecell{\textbf{Target-span}\\\textbf{pooling}} &
\makecell{\textbf{Target-cond.}\\\textbf{attention}} &
\makecell{\textbf{Char-}\\\textbf{CNN}} &
\makecell{\textbf{Orthographic}\\\textbf{features}} &
\makecell{\textbf{{$c$--$a$}}\\\textbf{interactions}} &
\makecell{\textbf{Gated}\\\textbf{discourse}} &
\makecell{\textbf{Training}\\\textbf{Macro-F1 (\%)}} &
\makecell{\textbf{Training}\\\textbf{Accuracy (\%)}} &
\makecell{\textbf{Validation}\\\textbf{Macro-F1 (\%)}} &
\makecell{\textbf{Validation}\\\textbf{Accuracy (\%)}} \\
\midrule

Raw BERTurk
& -- & -- & -- & -- & -- & -- & --
& 96.82\% & 96.93\% & 84.88\% & 85.55\% \\

BERTurk + \texttt{[TGT]}
& \checkmark & -- & -- & -- & -- & -- & --
& 98.15\% & 98.22\% & 85.59\% & 86.31\% \\

Target-Pooling BERTurk
& \checkmark & \checkmark & -- & -- & -- & -- & --
& 98.06\% & 98.13\% & 86.06\% & 86.58\% \\

Pool+Attention Hybrid
& \checkmark & \checkmark & \checkmark & -- & -- & -- & --
& 99.76\% & 99.77\% & 85.97\% & 86.51\% \\

OrthoTarget Hybrid CE
& \checkmark & \checkmark & \checkmark & \checkmark & \checkmark & -- & --
& \textbf{99.84\%} & \textbf{99.85\%} & 85.78\% & 86.38\% \\

\textbf{StutterStigma-BERT (Ours)}
& \checkmark & \checkmark & \checkmark & \checkmark & \checkmark &
\checkmark & \checkmark
& 99.83\% & 99.84\% &
\textbf{86.18\%} & \textbf{86.79\%} \\

\bottomrule
\end{tabular}
\end{adjustbox}

\vspace{1mm}
\begin{minipage}{0.98\textwidth}
\footnotesize
\textit{Note.}
Raw BERTurk uses preprocessed text without target markers and serves as the
contextual reference for the controlled comparison. Values are means across
three random-seed runs (42, 123, and 2026) under the common 20-epoch
training protocol. For each seed, training and validation metrics are measured
at the same checkpoint selected by the highest internal-validation Macro-F1.
Training metrics are reported descriptively, whereas architecture selection is
based on validation Macro-F1. Bold values indicate the highest mean value
within each reported metric. The $c$--$a$ interaction column denotes the
additional $|c-a|$ and $c\odot a$ terms in the expanded semantic
representation. CE = cross-entropy.
\end{minipage}
}
\end{table*}

\subsection{Model Performance}
\label{sec:model_performance}

Table~\ref{tab:final_model_performance} summarizes the predictive
performance of the validation-selected final StutterStigma-BERT
configurations across 20 random seeds evaluated on the same fixed held-out test partition. The models for both tasks are trained with cross-entropy loss function under the
common training protocol. The binary results additionally use a
decision threshold selected independently from the corresponding validation
split for each seed.

The primary three-class StutterStigma-BERT model achieves a mean Macro-F1 of 73.54\%,
with a mean Accuracy of 74.36\% and a mean Weighted-F1 of 74.45\%. In this model,
mean Macro-Precision and Macro-Recall are 73.68\% and 73.85\%,
respectively. 
The secondary binary StutterStigma-BERT model achieves a mean Macro-F1 of 85.17\%,
a mean Accuracy of 85.77\%, and a mean Weighted-F1 of 85.69\%. In this model, 
mean Macro-Precision and Macro-Recall are 85.71\% and 84.84\%,
respectively, with mean ROC-AUC of 87.83\% and mean PR-AUC of
84.68\%.
The higher aggregate scores in the binary task should not be interpreted
as a direct improvement over the three-class task because the two tasks
use different label spaces: the binary formulation merges
\textit{Non-stigmatizing/\allowbreak contextual} and \textit{Supportive} into a single
\textit{Non-stigma} category.

Fig.~\ref{fig:threeclass_final_learning_curves} presents the learning curves of the proposed three-class StutterStigma-BERT model, averaged across three random seeds. The model fits the training data rapidly: training loss continues to decrease throughout fine-tuning, and training Macro-F1 approaches saturation. In contrast, the mean validation loss reaches its minimum at epoch~4 and increases thereafter, whereas validation Macro-F1 remains comparatively stable after the initial improvement and attains its highest epoch-wise mean at epoch~15. This divergence between validation loss and validation Macro-F1 motivates the use of validation Macro-F1, rather than the terminal training epoch, as the checkpoint selection criterion. It should be noted that checkpoint selection is performed independently for each seed and the averaged curves serve only to characterize the overall training dynamics. A similar pattern is observed for the final binary configuration (Fig.~\ref{fig:binary_final_learning_curves}). Training loss decreases sharply, and training Macro-F1 approaches saturation. The mean validation loss again reaches its minimum at epoch~4, whereas validation Macro-F1 remains within a relatively narrow range after the early epochs and attains its highest epoch-wise mean at epoch~11. As in the three-class task, these mean curves are descriptive, and checkpoint selection is performed independently for each seed based on its highest validation Macro-F1.

The learning curves for both tasks reveal a clear train--validation divergence during continued fine-tuning: training loss continues to decrease and training Macro-F1 approaches saturation, while validation loss begins to increase after the early epochs. This behavior is consistent with prior work showing that high-capacity pre-trained language models, including BERT-based models, are susceptible to overfitting when fine-tuned on comparatively limited labelled downstream data \cite{jiang2020smart,wang2020integrating,somayajula2024generalizable}. Importantly, however, the increase in validation loss is not accompanied by a corresponding decline in validation Macro-F1. Such decoupling is well documented in modern neural networks: the cross-entropy loss continues to penalize increasingly overconfident predictions, particularly on misclassified samples, so that the loss can rise even when the predicted labels, and hence classification metrics, remain largely unchanged \cite{guo2017calibration}. Fine-tuned pre-trained language models have likewise been shown to suffer from miscalibration due to over-parameterization \cite{kong2020calibrated}. The rising validation loss therefore suggests increasing prediction confidence rather than a deterioration in discriminative performance. Moreover, BERT fine-tuning is known to be sensitive to optimization dynamics and random-seed effects \cite{dodge2020fine,mosbach2020stability}, which further supports selecting checkpoints according to validation performance rather than relying on the terminal training epoch. This strategy corresponds to the established practice of early stopping \cite{prechelt1998early}, with validation Macro-F1 adopted as the stopping criterion because it is the primary evaluation metric of this study and weights all classes equally.

A closely related social-media example is reported by Malik et al. \cite{malik2024categorization}. In their multiclass tweet-classification experiment with 128-token inputs, training loss decreased from 0.35 at epoch~1 to 0.09 at epoch~4, whereas validation loss reached its minimum of 0.25 at epoch~2 and subsequently increased to 0.38 by epoch~4 and the authors interpreted this increase as an indication that further training could lead to overfitting. Consistent with this observation and with standard early-stopping practice, the 20-epoch schedule in the present study is treated as a maximum training budget rather than a fixed training duration, and each run retains the checkpoint with the highest internal-validation Macro-F1. The held-out test set is never used for checkpoint selection or hyperparameter tuning, therefore, the reported test performance is not optimistically biased by test-set tuning. In addition, the alternative-backbone analyses reported in Section~\ref{sec:ablation_results} reveal substantial train--validation separation under the TurkishBERTweet and Turkish Mamba protocols as well, with larger gaps than those observed for the BERTurk-based model. Thus, although overfitting is evident in the present experiments, the observed train--validation divergence is not specific to the BERTurk-based implementation but reflects a general characteristic of fine-tuning high-capacity models on limited labeled data.

\subsection{Ablation Study on StutterStigma-BERT Model}
\label{sec:ablation_results}


\subsubsection{Ablation Study on Model Components for Three-Class Classification}
Table~\ref{tab:three_class_ablation} summarizes the results of three-seed ablation study on the model components of three-class classification scheme evaluated under the common training protocol to show the effectiveness of each component in our final proposed StutterStigma-BERT model. As presented in this table, raw BERTurk reaches a mean validation
Macro-F1 of 75.01\%. Explicit target marking (BERTurk+\texttt{[TGT]}) increases this value to
75.37\%, and Target-Pooling module further increases it to 75.81\%.
Although adding target-conditioned attention alone slightly decreases the
mean validation Macro-F1 from 75.81\% to 75.53\%, the attention
representation is retained because it provides target-conditioned contextual
information that is subsequently combined with the orthographic features,
$c$--$a$ interactions, and gated discourse representation. With these
components combined, the final StutterStigma-BERT model reaches the highest
mean validation Macro-F1 of 76.32\% and a validation Accuracy of 77.18\%,
and is therefore retained as our final architecture.
The transition from
OrthoTarget to the complete gated configuration introduces both the additional
$c$--$a$ semantic interactions and the gated discourse representation;
their contribution is therefore interpreted jointly at the configuration level.

\subsubsection{Ablation Study on Model Components for Binary Classification}

Table~\ref{tab:binary_ablation} summarizes the results of three-seed ablation study on the model components of binary classification scheme evaluated under the common training protocol to show the effectiveness of each component in our final proposed  StutterStigma-BERT model. For the binary task, 
Raw BERTurk reaches a mean
validation Macro-F1 of 84.88\%. Explicit target marking increases this value
to 85.59\%, while Target-Pooling reaches 86.06\%. 
Although the intermediate additions do not produce monotonic improvements,
combining target-conditioned attention with the OrthoTarget features,
$c$--$a$ interactions, and gated discourse representation produces the
highest-performing complete configuration. Our final StutterStigma-BERT
model therefore reaches the highest mean validation Macro-F1 of 86.18\%
and a validation Accuracy of 86.79\%.
Macro-F1 remains the primary criterion for architecture selection because it
weights the two classes equally and is less influenced by differences in class
frequency than overall Accuracy.


\begin{table*}[t]
\caption{Backbone Ablation of the Proposed StutterStigma-BERT Model (BERTurk, TurkishBERTweet, and Turkish Mamba) Across Model Components (Mean Over Three Seeds)}
\label{tab:backbone_ablation_comparison}
\centering
\scriptsize
\setlength{\tabcolsep}{3.0pt}
\renewcommand{\arraystretch}{1.13}
{

\begin{adjustbox}{max width=\textwidth}
\begin{tabular}{lccc|ccc|ccc}
\toprule
&
\multicolumn{3}{c|}{\textbf{BERTurk}} &
\multicolumn{3}{c|}{\textbf{TurkishBERTweet}} &
\multicolumn{3}{c}{\textbf{Turkish Mamba}} \\
\cmidrule(lr){2-4}
\cmidrule(lr){5-7}
\cmidrule(lr){8-10}

\textbf{Model Components}
&
\makecell{\textbf{Training}\\\textbf{Macro-F1 (\%)}}
&
\makecell{\textbf{Validation}\\\textbf{Macro-F1 (\%)}}
&
\makecell{\textbf{Gap}\\\textbf{(pp)}}
&
\makecell{\textbf{Training}\\\textbf{Macro-F1 (\%)}}
&
\makecell{\textbf{Validation}\\\textbf{Macro-F1 (\%)}}
&
\makecell{\textbf{Gap}\\\textbf{(pp)}}
&
\makecell{\textbf{Training}\\\textbf{Macro-F1 (\%)}}
&
\makecell{\textbf{Validation}\\\textbf{Macro-F1 (\%)}}
&
\makecell{\textbf{Gap}\\\textbf{(pp)}} \\
\midrule

Plain backbone
& 94.40\% & 75.01\% & 19.39
& 96.68\% & 70.77\% & 25.91
& 99.83\% & 66.03\% & 33.80 \\

+\texttt{[TGT]}
& 92.51\% & 75.37\% & 17.14
& 99.33\% & 69.87\% & 29.46
& 99.83\% & 66.04\% & 33.79 \\

Target-Pooling
& 99.15\% & 75.81\% & 23.34
& 97.64\% & 71.24\% & 26.40
& 99.92\% & 65.75\% & 34.17 \\

Pool+Attention
& 96.87\% & 75.53\% & 21.34
& 99.25\% & \textbf{71.25\%} & 28.00
& 99.97\% & \textbf{67.73\%} & 32.24 \\

OrthoTarget CE
& 97.29\% & 76.05\% & 21.24
& 95.99\% & 71.00\% & 24.99
& 96.96\% & 66.45\% & 30.51 \\

\textbf{All Components in StutterStigma-BERT (Ours)
}
& 99.13\% & \textbf{76.32\%} & 22.81
& 97.86\% & 70.41\% & 27.45
& 97.51\% & 66.63\% & 30.88 \\

\bottomrule
\end{tabular}
\end{adjustbox}

\vspace{1mm}
\begin{minipage}{0.98\textwidth}
\footnotesize
\textit{Note.}
Values are mean Macro-F1 percentages across the same three random-seed runs
(42, 123, and 2026), measured at each run's validation-selected checkpoint.
Gap denotes mean training Macro-F1 minus mean internal-validation Macro-F1
in percentage points (pp). The six configuration levels are functionally
matched across backbones, with backbone-specific adaptations described in
this section. The BERTurk implementation of the proposed architecture corresponds to the
proposed three-class StutterStigma-BERT model. Bold validation values indicate the highest mean
internal-validation Macro-F1 within each backbone. Gap values are reported
descriptively and are not used for model selection.
\end{minipage}
}
\end{table*}

\subsubsection{Ablation Study on Different Backbones for Three-class Classification}

The alternative-backbone ablation study examines whether the same target-aware
configuration progression in our proposed model retains similar behavior when the BERTurk encoder
is replaced with other pretrained language-model backbones. Two alternative
backbones are evaluated for the primary three-class task: Turkish Mamba
\cite{gu2023mamba} and TurkishBERTweet
\cite{najafi2024turkishbertweet}. For each backbone, the same six-stage
functional progression is examined: a plain backbone classifier, explicit
target marking, Target-Pooling, Pool+Attention, OrthoTarget, and +gated discourse (all components together) and the results are presented in Table \ref{tab:backbone_ablation_comparison}. Configuration comparisons are based on mean
internal-validation Macro-F1 across seeds 42, 123, and 2026, using the
validation-best checkpoint from each run.

For Mamba backbone, we replace BERTurk with the
\texttt{serda-dev/mamba-370m-hf-turkish} checkpoint, based on the Mamba
state-space architecture \cite{gu2023mamba}. All experiments based on Mamba backbone use
the common data split, three random seeds, 20-epoch budget, optimizer,
learning rate, warm-up ratio, weight decay, task dropout, label smoothing,
R-Drop coefficient, batch sizes, and a maximum sequence length of 160 tokens
wherever these settings are directly applicable. The character-level,
handcrafted orthographic, and discourse representations retain the same
functional definitions as in the corresponding StutterStigma-BERT
configurations.
Since Mamba does not use a BERT-style \texttt{[CLS]} representation, the
final hidden state of the last non-padding token is used as the global post
representation. To preserve the same backbone-freezing proportion, the
bottom 16 of 48 Mamba blocks are frozen, corresponding to the bottom four
of 12 BERTurk encoder blocks. BERT-specific hidden-state and
attention-probability dropout have no direct Mamba equivalent and are
therefore not imposed on the Mamba backbone. For the target-aware Mamba
configurations, parameter-free layer normalization is applied only to copies
of the $c$, $t$, and $a$ representations used in the pairwise interaction
terms and discourse-gate conditioning input. No additional learned
parameters are introduced by this normalization.

The second backbone in this ablation study is
\texttt{VRLLab/ TurkishBERTweet}, a RoBERTa-based language model pretrained
on large-scale Turkish social-media data
\cite{najafi2024turkishbertweet}. All experiments based on TurkishBERTweet use the same fixed data
split, three random seeds, 20-epoch training budget, optimizer, learning
rate, warm-up ratio, weight decay, task dropout, label smoothing, R-Drop
coefficient, batch sizes, and validation-based checkpoint-selection
procedure as the common controlled protocol. The bottom four of the
12 Transformer blocks are frozen, while the embedding layer and upper
eight blocks remain trainable. For target-aware configurations, the hidden
state of the first sequence token is used as the global representation $c$.
The target-span, target-conditioned attention, character-level,
orthographic, and discourse components follow the corresponding
StutterStigma-BERT definitions.
TurkishBERTweet is evaluated using a maximum sequence length of 128 tokens,
following the setting used in the original TurkishBERTweet study
\cite{najafi2024turkishbertweet}. For target-marked inputs, ordinary right
truncation is used when both target-boundary markers remain within the
sequence. When truncation would remove either \texttt{[TGT]} or
\texttt{[/TGT]}, a deterministic target-preserving rule retains the complete
marked target span and allocates the remaining token budget to the nearest
available left and right context. This rule is independent of the class label.

As presented in Table \ref{tab:backbone_ablation_comparison}, the BERTurk 
backbone consistently achieves the highest performance across all component 
configurations compared with the two other backbones. When all proposed 
components are integrated, the BERTurk-based model attains the highest 
validation Macro-F1 of 76.32\%, compared with 70.41\% for TurkishBERTweet 
and 66.63\% for Mamba. Even in their best-performing configuration 
(Pool+Attention), TurkishBERTweet and Mamba reach validation Macro-F1 
scores of only 71.25\% and 67.73\%, respectively, both of which remain 
considerably lower than the 75.53\% achieved by the BERTurk-based model 
under the same configuration. Overall, BERTurk yields the best performance, 
followed by TurkishBERTweet and Mamba. Accordingly, BERTurk is selected 
as the backbone of the proposed StutterStigma-BERT model. 

Furthermore, the 
gap between training and validation Macro-F1 scores in Table 
\ref{tab:backbone_ablation_comparison} indicates that, although all three 
backbones exhibit some degree of overfitting, it is notably less in the BERTurk-
based StutterStigma-BERT model (22.81~pp) than in the TurkishBERTweet- and Mamba-
based models (27.45~pp and 30.88~pp, respectively). A gap between training and validation 
performance is commonly observed when fine-tuning large pre-trained 
language models on downstream tasks with limited labelled data. Owing to 
their very high capacity (approximately 110M parameters for BERT-base 
models such as BERTurk \cite{devlin2019bert}), these models can fit the 
training set almost perfectly, and fine-tuning them on small datasets is known 
to cause overfitting to the downstream data \cite{howard2018ulmfit, 
jiang2020smart,malik2024categorization}. However, over-parameterized deep 
networks can fit their training data closely while still generalizing well 
\cite{zhang2016understanding}, hence, generalization is more appropriately 
assessed by performance on held-out data than by the training--validation gap 
alone. In this respect, the proposed StutterStigma-BERT achieves the highest 
validation Macro-F1 among all evaluated backbones while also exhibiting the 
smallest training--validation gap, which is consistent with the better held-out generalization of the BERTurk-based model among the evaluated backbones. In addition, 
regularization strategies, including partial encoder
freezing, label smoothing, R-Drop consistency regularization, and
validation-based checkpoint selection alongside dropout and weight decay have been 
employed to mitigate overfitting. Additional evidence from low-resource fine-tuning studies similarly indicates
that pretrained language models can remain susceptible to overfitting and
training instability when task-specific labeled data are limited
\cite{wang2020integrating,somayajula2024generalizable}. 

\begin{table}[!t]
\caption{Held-Out Test Performance of the Final StutterStigma-BERT Model for Each of the 20 Random Seeds (Three-Class and Binary Tasks)}
\label{tab:final_20seed_robustness}
\centering
\scriptsize
\setlength{\tabcolsep}{7pt}
\renewcommand{\arraystretch}{1.08}
{
\begin{tabular}{c cc cc c}
\toprule
\textbf{Seed} &
\multicolumn{2}{c}{\textbf{Three-class}} &
\multicolumn{3}{c}{\textbf{Binary}} \\
\cmidrule(lr){2-3}
\cmidrule(lr){4-6}
&
\makecell{\textbf{Accuracy}\\\textbf{(\%)}} &
\makecell{\textbf{Macro-F1}\\\textbf{(\%)}} &
\makecell{\textbf{Accuracy}\\\textbf{(\%)}} &
\makecell{\textbf{Macro-F1}\\\textbf{(\%)}} &
\makecell{\textbf{Selected}\\\textbf{threshold}} \\
\midrule
7    & 74.09\% & 73.52\% & 85.30\% & 84.84\% & 0.490 \\
42   & 74.50\% & 73.85\% & 86.13\% & 85.60\% & 0.085 \\
123  & 74.42\% & 73.93\% & 86.13\% & 85.45\% & 0.130 \\
269  & 74.50\% & 73.72\% & 86.13\% & 85.42\% & 0.480 \\
317  & 72.76\% & 72.47\% & 85.71\% & 85.09\% & 0.730 \\
421  & 74.75\% & 74.15\% & 86.05\% & 85.26\% & 0.300 \\
2026 & 74.83\% & 73.80\% & 85.88\% & 85.14\% & 0.625 \\
2208 & 73.50\% & 72.50\% & 85.55\% & 84.89\% & 0.380 \\
3314 & 74.58\% & 73.74\% & 85.55\% & 84.88\% & 0.370 \\
3359 & 75.25\% & 73.93\% & 84.72\% & 84.17\% & 0.125 \\
3407 & 74.75\% & 73.86\% & 85.80\% & 85.21\% & 0.475 \\
3801 & 75.17\% & 73.82\% & 85.96\% & 85.39\% & 0.780 \\
5535 & 74.67\% & 74.07\% & 86.30\% & 85.78\% & 0.350 \\
5858 & 74.58\% & 73.95\% & 85.63\% & 85.02\% & 0.375 \\
6022 & 74.50\% & 73.29\% & 85.71\% & 85.21\% & 0.085 \\
7002 & 74.50\% & 74.09\% & 85.55\% & 85.06\% & 0.155 \\
9234 & 74.42\% & 73.16\% & 85.88\% & 85.36\% & 0.075 \\
9251 & 74.83\% & 73.68\% & 86.38\% & 85.83\% & 0.545 \\
9820 & 73.84\% & 73.05\% & 84.88\% & 84.36\% & 0.050 \\
9901 & 72.84\% & 72.23\% & 86.21\% & 85.47\% & 0.605 \\
\midrule
\textbf{Mean} &
\textbf{74.36\%} &
\textbf{73.54\%} &
\textbf{85.77\%} &
\textbf{85.17\%} &
\textbf{0.3605} \\
\bottomrule
\end{tabular}

\vspace{1mm}
\begin{minipage}{0.49\textwidth}
\footnotesize
\textit{Note.}
Both tasks use our final proposed StutterStigma-BERT model
trained with cross-entropy and the
same fixed held-out test set of 1,204 posts.
Values are shown separately for each of the 20 distinct random seeds.
For the binary task, the decision threshold is selected independently from
the corresponding validation split and subsequently applied unchanged to
the held-out test probabilities. 
\end{minipage}
}
\end{table}

\begin{table*}[t]
\caption{
Decision-Threshold Sensitivity of the Final Binary
StutterStigma-BERT Model Across 20 Random Seeds on Held-out Test Data
}
\label{tab:binary_threshold_sensitivity}
\centering
\scriptsize
\setlength{\tabcolsep}{4pt}
\renewcommand{\arraystretch}{1.10}

{
\begin{tabular}{lcccccc}
\toprule
\textbf{Threshold policy} &
\makecell{\textbf{Threshold}\\\textbf{mean}} &
\makecell{\textbf{Accuracy}\\\textbf{mean (\%)}} &
\makecell{\textbf{Macro-F1}\\\textbf{mean (\%)}} &
\makecell{\textbf{Stigmatizing}\\\textbf{recall, mean (\%)}} &
\textbf{Mean FN} &
\textbf{Mean FP} \\
\midrule

Default 0.50 &
0.5000 &
85.84\% &
85.17\% &
78.17\% &
108.70 &
61.80 \\

Validation Macro-F1 (applied in our final StutterStigma-BERT model) &
0.3605 &
85.77\% &
85.17\% &
79.43\% &
102.45 &
68.85 \\

\bottomrule
\end{tabular}

\vspace{1mm}
\begin{minipage}{0.98\textwidth}
\footnotesize
\textit{Note.} Values summarize the same 20 random-seed evaluations of the proposed StutterStigma-BERT model
for binary classification on the held-out test data. The validation Macro-F1
threshold is selected independently within each corresponding validation
split and then applied unchanged to the fixed held-out test set. The default
policy uses a fixed threshold of 0.50. FN = false negatives; FP = false
positives.
\end{minipage}
}
\end{table*}

\subsection{Final Robustness Analysis Across Random Seeds}
\label{sec:final_robustness}


The validation-selected final StutterStigma-BERT model is evaluated across
20 distinct random seeds for both the three-class and binary tasks on the
fixed held-out test set. The 20-seed set includes the three seeds used in
the controlled ablation analysis together with 17 additional random seeds.
These repeated evaluations quantify sensitivity to random initialization
and training stochasticity and do not constitute cross-validation or
independent external validation. Results for each seed are presented in
Table~\ref{tab:final_20seed_robustness} based on Accuracy and Macro-F1.
For each seed, the model is trained for a maximum of 20 epochs and the
checkpoint with the highest internal-validation Macro-F1 is retained.
The corresponding checkpoint is then evaluated once on the same fixed
held-out test set of 1,204 posts. For the binary task, the decision threshold
is selected independently from the corresponding validation split and
applied unchanged to the held-out test probabilities.

Repeated-seed robustness is examined using paired
prediction-level statistical comparisons because all seed-specific models
are evaluated on the same fixed set of 1,204 held-out test examples.
For each task, Cochran's $Q$ test is first used as an omnibus test of
differences in per-example correctness across the 20 seed-specific model
instances. All $\binom{20}{2}=190$ seed pairs are then compared
individually. Pairwise differences in Accuracy are evaluated using the
exact McNemar test based on discordant correctness outcomes, whereas
pairwise differences in Macro-F1 are evaluated using a paired Monte Carlo
randomization test with 20,000 within-example exchanges of the two
seed-specific predictions. Family-wise error arising from the 190
comparisons is controlled separately for Accuracy and Macro-F1 within
each task using the Holm correction. Statistical significance is evaluated
at $\alpha=0.05$, and pairwise significance is interpreted using the
Holm-adjusted $p$ values.


As presented in Table~\ref{tab:final_20seed_robustness}, 
across the 20 seeds on held-out test data, the final three-class model achieves a mean held-out
Accuracy of 74.36\% and a mean Macro-F1 of 73.54\%
(95\% CI: 73.27\%--73.81\%). Additionally, mean Weighted-F1 is 74.45\%.
Class-wise F1 is highest for \textit{Stigmatizing} posts (81.61\%),
followed by \textit{Supportive} posts (73.90\%), whereas
\textit{Non-stigmatizing/contextual} remains the most difficult class
(65.10\%). 
At $\alpha = 0.05$, Cochran's Q test does not indicate an overall difference
in per-example correctness across the 20 seed-specific three-class models,
$Q(19)=19.83$, $p=.405$. Across the 190 pairwise seed comparisons,
15 Accuracy comparisons have unadjusted $p<.05$ under the exact
McNemar test and six Macro-F1 comparisons have unadjusted $p<.05$
under the paired randomization test. However, none of these differences
remain statistically significant after Holm correction
(0/190 for Accuracy and 0/190 for Macro-F1), showing the robustness of the model performance over 20 seeds.


Across the same 20 random seeds on the held-out test data, the final binary model achieves a mean
held-out Accuracy of 85.77\% and a mean Macro-F1 of 85.17\%
(95\% CI: 84.98\%--85.37\%). Additionally, mean Weighted-F1 is 85.69\%.
For the \textit{Stigmatizing} class, mean 
F1 is
82.20\%, while the
\textit{Non-stigma} class reaches a mean F1 of 88.15\%.
Mean ROC-AUC and PR-AUC are 87.83\% and 84.68\%, respectively.
The mean validation-selected decision threshold across runs is 0.3605. Cochran's $Q$ test likewise does not indicate an overall difference in
per-example correctness across the 20 seed-specific binary models,
$Q(19)=14.18$, $p=.773$. Among the 190 pairwise seed comparisons,
one Accuracy comparison has an unadjusted $p<.05$ under the exact
McNemar test and one Macro-F1 comparison has an unadjusted $p<.05$
under the paired randomization test. Neither difference remains
statistically significant after Holm correction
(0/190 for Accuracy and 0/190 for Macro-F1).
Taken together, the repeated-seed analyses indicate that the held-out
performance of the final configuration is stable across the evaluated
random seeds in both classification tasks. Within the evaluated seed set, the observed variation therefore provides no
evidence of systematic seed-dependent performance differences.


\begin{table*}[t]
\caption{Comparison of the Proposed StutterStigma-BERT Model with Lexical and Contextual Baselines on the Held-Out Test Set}
\label{tab:main_baseline_comparison}
\centering
\small
\setlength{\tabcolsep}{5pt}
\renewcommand{\arraystretch}{1.10}

{
\begin{adjustbox}{max width=\textwidth}
\begin{tabular}{lllccc}
\toprule
\textbf{Task} &
\textbf{Model} &
\textbf{Evaluation setting} &
\textbf{Accuracy (\%)} &
\textbf{Macro-F1 (\%)} &
\textbf{Role} \\
\midrule

Three-class &
TF--IDF + Logistic Regression \cite{salton1988term} &
Single fit &
65.28\% &
63.82\% &
Lexical baseline \\

Three-class &
TF--IDF + Linear SVM \cite{salton1988term} &
Single fit &
66.36\% &
64.81\% &
Lexical baseline \\

Three-class &
Raw BERTurk \cite{schweter_2020_berturk} &
Common three-seed mean &
73.23\% &
72.48\% &
Contextual baseline \\

Three-class &
\textbf{StutterStigma-BERT (Ours)} &
Common three-seed mean &
\textbf{74.92\%} &
\textbf{74.19\%} &
Proposed model \\

\midrule

Binary &
TF--IDF + Logistic Regression \cite{salton1988term} &
Single fit &
78.57\% &
77.93\% &
Lexical baseline \\

Binary &
TF--IDF + Linear SVM \cite{salton1988term} &
Single fit &
79.32\% &
78.04\% &
Lexical baseline \\

Binary &
Raw BERTurk \cite{schweter_2020_berturk} &
Common three-seed mean, validation threshold &
85.77\% &
85.28\% &
Contextual baseline \\

Binary &
\textbf{StutterStigma-BERT (Ours)} &
Common three-seed mean, validation threshold &
\textbf{86.05\%} &
\textbf{85.40\%} &
Proposed model \\

\bottomrule
\end{tabular}
\end{adjustbox}

\vspace{1mm}
\begin{minipage}{0.98\textwidth}
\footnotesize
\textit{Note.} TF--IDF models are deterministic single-fit lexical
baselines. Raw BERTurk and the proposed StutterStigma-BERT configurations
are reported as means across the same three random seeds
(42, 123, and 2026) under the common 20-epoch neural training protocol.
Binary neural metrics use the decision threshold selected independently
from the corresponding validation split. Final 20-seed performance is reported separately in
Table~\ref{tab:final_model_performance}. CE = cross-entropy.
\end{minipage}
}
\end{table*}

\subsection{Binary Decision-Threshold Sensitivity}
\label{sec:binary_threshold_sensitivity}

For the secondary binary task, decision-threshold selection is performed separately for each random-seed run and the results over 20 seeds are presented in Table \ref{tab:binary_threshold_sensitivity}. The validation-best checkpoint first produces probabilities for the positive \textit{Stigmatizing} class on the corresponding validation subset. Candidate thresholds are then evaluated over the range 0.05--0.95 in increments of 0.005 as follows:

\begin{equation}
\begin{aligned}
\theta^{*} &= \arg\max_{\theta \in \Theta} \operatorname{MacroF1}_{\mathrm{val}}(\theta),\\
\Theta &= \{0.050, 0.055, \ldots, 0.950\}.
\end{aligned}
\label{eq:validation_threshold_selection}
\end{equation}

When multiple thresholds produce the same maximum validation Macro-F1, the lowest threshold is retained. The selected threshold is then frozen and applied unchanged to the held-out test probabilities produced by the corresponding seed-specific checkpoint \cite{lipton2014thresholding}. Threshold selection changes only the probability-to-label decision rule and it
does not modify the learned model parameters. It should therefore not be
interpreted as probability calibration.

Our proposed StutterStigma-BERT model for binary classification uses a
seed-specific decision threshold selected exclusively from the corresponding
validation split by maximizing validation Macro-F1. As presented in Table \ref{tab:binary_threshold_sensitivity}, across the 20 random
seeds, the mean validation-selected threshold is $0.3605 \pm 0.2300$.
To characterize the effect of this decision rule, the resulting test-set
predictions are compared with those obtained from the trained
models over the same 20 seeds using the fixed default threshold of 0.50. No threshold is selected
or modified using the held-out test labels.
The two threshold policies yield essentially identical mean test Macro-F1
of 85.17\%. The validation-selected policy increases mean
\textit{Stigmatizing} recall from 78.17\% to 79.43\% and reduces the
mean number of false negatives from 108.70 to 102.45. This increase in
recall is accompanied by an increase in mean false positives from 61.80 to
68.85, while mean Accuracy changes only slightly from 85.84\% to 85.77\%.
Thus, validation-based threshold selection primarily changes the balance
between missed stigmatizing posts and false alarms rather than overall
Macro-F1.

Threshold selection is performed after the model architecture and
validation-best checkpoint have been determined and therefore does not
affect architecture selection or learned model parameters. The substantial
variation in the selected thresholds across seeds further indicates that the
probability operating point is sensitive to training stochasticity even when
the architecture and optimization protocol are fixed.

\subsection{
Performance Comparison with Baselines and Existing Approaches in the Literature}
\label{sec:baseline_comparisons}

The proposed StutterStigma-BERT model is compared with three complementary groups of
reference models for both three-class and binary classification tasks. First, it is compared with lexical TF--IDF models and Raw BERTurk models, providing
within-task baselines. Second, it is compared with existing task-adapted neural comparators in the literature (harmful-language
classifiers) by training them on our collected
StutterStigma-TR dataset for 20 epochs under the common training and evaluation protocol to provide
a controlled in-domain comparison. Third, these existing Turkish harmful-language
classifiers are evaluated separately by direct transfer, without directly
training on StutterStigma-TR dataset, to examine cross-task transfer from broader
harmful-language detection. These comparisons address different questions
and are therefore interpreted separately.

\subsubsection{Comparison with the Lexical and Contextual Baselines}
Table~\ref{tab:main_baseline_comparison} compares the performance of the final StutterStigma-BERT model with three lexical and contextual baselines including TF--IDF + Logistic Regression \cite{salton1988term}, TF--IDF + Linear SVM \cite{salton1988term}, and Raw BERTurk \cite{schweter_2020_berturk}, on the held-out test data, over three seeds, for both three-class and binary classification tasks.
The lexical baselines use word-level TF--IDF features with unigram and
bigram features, a minimum document frequency of 2, a maximum vocabulary
size of 50,000 features, and sublinear term-frequency scaling. Logistic
Regression uses balanced class weights, the \texttt{liblinear} solver, and a
maximum of 3,000 iterations. The Linear SVM also uses balanced class weights
and is calibrated using sigmoid calibration with three-fold cross-validation.
Both lexical models are fitted once on the fixed training partition using
\texttt{preprocessed\_text}.

\begin{table*}[t]
\caption{Comparison of the Proposed StutterStigma-BERT Model with Existing Turkish Harmful-Language Models Fine-Tuned on StutterStigma-TR (Mean Over Three Seeds)}
\label{tab:task_adapted_comparators}
\centering
\small
\setlength{\tabcolsep}{4.5pt}
\renewcommand{\arraystretch}{1.10}

{
\begin{tabular}{llccc}
\toprule
\textbf{Task} &
\textbf{Model} &
\makecell{\textbf{Accuracy}\\\textbf{mean (\%)}} &
\makecell{\textbf{Macro-F1}\\\textbf{mean (\%)}} &
\makecell{\textbf{Weighted-F1}\\\textbf{mean (\%)}} \\
\midrule

Three-class &
KUISAIL-style BERT--CNN \cite{safaya2020kuisail} &
73.48\% &
72.44\% &
73.45\% \\

Three-class &
Hate-speech-pretrained BERTurk \cite{toraman2022large} &
71.95\% &
71.15\% &
72.13\% \\

Three-class &
Bullying-pretrained BERTurk \cite{sarigil2022bullyingmodel} &
72.29\% &
71.54\% &
72.44\% \\

Three-class &
Toxicity-pretrained ConvBERT \cite{uludogan2023toxicitymodel} &
68.74\% &
68.19\% &
68.99\% \\

Three-class &
\textbf{StutterStigma-BERT (Ours)
} &
\textbf{74.92\%} &
\textbf{74.19\%} &
\textbf{75.05\%} \\

\midrule

Binary &
KUISAIL-style BERT--CNN \cite{safaya2020kuisail} &
84.97\% &
84.43\% &
84.91\% \\

Binary &
Hate-speech-pretrained BERTurk \cite{toraman2022large} &
83.33\% &
82.72\% &
83.28\% \\

Binary &
Bullying-pretrained BERTurk \cite{sarigil2022bullyingmodel} &
84.55\% &
83.70\% &
84.34\% \\

Binary &
Toxicity-pretrained ConvBERT \cite{uludogan2023toxicitymodel} &
79.96\% &
79.18\% &
79.86\% \\

Binary &
\textbf{StutterStigma-BERT (Ours)} &
\textbf{86.05\%} &
\textbf{85.40\%} &
\textbf{85.93\%} \\

\bottomrule
\end{tabular}

\vspace{1mm}
\begin{minipage}{0.98\textwidth}
\footnotesize
\textit{Note.} Values are mean percentages across the same three
random-seed runs (42, 123, and 2026). All models are trained on
StutterStigma-TR using the same fixed train--test partition and the common
20-epoch model-agnostic hyperparameter and validation-based model-selection
protocol, where architecture-applicable. The proposed-model rows report
the corresponding three-seed common-protocol values rather than the
separate 20-seed robustness estimates. Binary metrics use the decision
threshold selected independently from the corresponding validation split. Bold values indicate the best results.
\end{minipage}
}

\end{table*}

As presented in Table~\ref{tab:main_baseline_comparison}, the contextual Raw BERTurk baseline is substantially stronger than the lexical
TF--IDF baselines in both tasks. Under the matched three-seed neural protocol,
the three-class Macro-F1 increases from 72.48\% with Raw BERTurk to 74.19\%
with the proposed StutterStigma-BERT model. In the binary task,
however, raw BERTurk is already highly competitive and Macro-F1 slightly improves 
from 85.28\% (raw BERTurk) to 85.40\% (StutterStigma-BERT). In this matched three-seed held-out comparison,
the descriptive improvement over Raw BERTurk is larger for the three-class
task than for the binary task. Because these comparisons use three seeds and no significance test, the small binary difference (0.12~pp) should not be interpreted as evidence of a reliable improvement.

\subsubsection{Comparison with the Existing Harmful-Language Classifiers}

To provide a controlled comparison with the existing models in the literature developed for broader
Turkish harmful-language processing, Table~\ref{tab:task_adapted_comparators} compares the performance of the final StutterStigma-BERT model with four non-target-aware approaches: a KUISAIL-style BERT--CNN
\cite{safaya2020kuisail}, a hate-speech-pretrained BERTurk encoder
\cite{toraman2022large}, a bullying-pretrained BERTurk encoder
\cite{sarigil2022bullyingmodel}, and a toxicity-pretrained ConvBERT encoder
\cite{uludogan2023toxicitymodel}. Each model is evaluated for both 
three-class and binary tasks across the same three random seeds
(42, 123, and 2026).
For the three harmful-language-pretrained models, the original classification
heads are discarded and the pretrained checkpoints are used as encoder
initializations. A newly initialized StutterStigma-TR task head is then trained
for each classification setting. The KUISAIL-style model retains its
BERT--CNN architecture, using the final four transformer hidden layers as input
channels to convolutional filters with kernel heights 1--5, followed by
max-over-time pooling and a newly initialized task classifier.
All four models are trained for 20 epochs using the same train--test samples of our collected StutterStigma-TR dataset with the same hyperparameter and
validation-based model-selection protocol described in
Section~\ref{sec:experiment_setup}. They use the same epoch budget, seed set,
learning rate, regularization settings, batch sizes, and validation-best
Macro-F1 checkpoint criterion wherever these settings are applicable to the
corresponding architecture. For the binary task, the decision threshold is
selected independently from the corresponding validation split.
The task-adapted models receive unmarked post text and do not use the
\texttt{[TGT]} and \texttt{[/TGT]} boundary markers. Explicit target marking
and the associated target-aware representations are components of the proposed
StutterStigma-BERT framework and are therefore not added to the compared architectures.

As presented in Table~\ref{tab:task_adapted_comparators}, under the common three-seed evaluation protocol, the proposed
StutterStigma-BERT model obtains the highest mean
Macro-F1 among the compared task-adapted models in both three-class and binary classification tasks.
For three-class classification, its mean Macro-F1 is 74.19\%, compared with
72.44\% for the KUISAIL-style BERT--CNN, 71.54\% for the
bullying-pretrained BERTurk, 71.15\% for the hate-speech-pretrained BERTurk,
and 68.19\% for the toxicity-pretrained ConvBERT. For binary classification,
the corresponding values are 85.40\%, 84.43\%, 83.70\%, 82.72\%, and
79.18\%.


In a separate cross-task transfer analysis, the same three publicly available
Turkish harmful-language classifiers are evaluated without task-specific
adaptation. Unlike the controlled comparison above, these models are
directly evaluated on the fixed StutterStigma-TR held-out test set without training on the
StutterStigma-TR training set. This analysis therefore examines how
well decision functions learned for broader hate-speech, bullying, and
toxicity tasks transfer to stuttering-related stigma without in-domain
supervision.
The three models cover different forms of harmful language:
hate speech, bullying, and toxicity. The hate-speech model is
\path{ctoraman/hate-speech-berturk} \cite{toraman2022large},
the bullying model is
\path{nanelimon/bert-base-turkish-bullying}
\cite{sarigil2022bullyingmodel}, and the toxicity model is
\path{gokceuludogan/convbert-base-turkish-mc4-toxicity-uncased}
\cite{uludogan2023toxicitymodel}. For the bullying model, the
sexist bullying, racism, and teasing/insult labels are mapped to
the harmful class, while the neutral label is mapped to Non-stigma.

\begin{table*}[t]
\caption{Direct-Transfer Performance of Existing Turkish Harmful-Language Classifiers on the Binary Task (Without Training on StutterStigma-TR)}
\label{tab:direct_transfer_harmful}
\centering
\scriptsize
\setlength{\tabcolsep}{4pt}

\begin{tabular}{lccccccc}
\toprule
\textbf{Source model / task} &
\textbf{Accuracy (\%)} &
\textbf{Macro-F1 (\%)} &
\textbf{95\% CI} &
\makecell{\textbf{Stigma}\\\textbf{Precision (\%)}} &
\makecell{\textbf{Stigma}\\\textbf{Recall (\%)}} &
\makecell{\textbf{Stigma}\\\textbf{F1 (\%)}} &
\makecell{\textbf{ROC-AUC /}\\\textbf{PR-AUC (\%)}} \\
\midrule

Hate-speech BERTurk \cite{toraman2022large} &
68.27\% &
68.21\% &
65.45--70.82\% &
58.87\% &
77.31\% &
66.84\% &
75.81\% / 68.62\% \\

Bullying BERTurk \cite{sarigil2022bullyingmodel} &
69.35\% &
67.42\% &
64.72--70.08\% &
65.62\% &
54.42\% &
59.50\% &
73.42\% / 67.28\% \\

Toxicity ConvBERT \cite{uludogan2023toxicitymodel} &
65.78\% &
64.02\% &
61.19--66.71\% &
59.77\% &
52.81\% &
56.08\% &
70.03\% / 60.43\% \\

\bottomrule
\end{tabular}

\vspace{1mm}
\begin{minipage}{0.98\textwidth}
\footnotesize
\textit{Note.} The three models are evaluated by direct transfer, without
training on StutterStigma-TR. Their native labels are mapped to the binary
task. Values are reported as percentages. Confidence intervals are bootstrap
95\% intervals for Macro-F1 based on 2,000 bootstrap resamples of the fixed
held-out test set.
\end{minipage}
\end{table*}

As presented in Table \ref{tab:direct_transfer_harmful}, the direct-transfer results are substantially lower than the scores obtained
after in-domain task adaptation (presented in Table \ref{tab:task_adapted_comparators}). The hate-speech classifier reaches substantially lower
Macro-F1 of 68.21\%, followed by the bullying model at 67.42\% and the
toxicity model at 64.02\%. Because these models are evaluated without being trained on
StutterStigma-TR training data, the comparison should not be interpreted as a
controlled architecture comparison. Instead, the comparison between Tables \ref{tab:task_adapted_comparators} and \ref{tab:direct_transfer_harmful} indicates that decision
boundaries learned for broader harmful-language tasks do not transfer
directly to the condition-specific distinction examined here. The label spaces differ, and many posts in StutterStigma-TR depend on stance, reported stigma, supportive responses, or condition-specific stereotypes rather than broad toxicity cues. This supports the need for condition-specific labeled data such as StutterStigma-TR.

\subsection{Comparison with the Zero-Shot Large Language Models}
\label{sec:zero_shot_llm}

We also evaluate two instruction-tuned large language models in a
zero-shot setting: \path{Qwen/Qwen2.5-7B-Instruct}
\cite{qwen2024qwen25} and
\path{ytu-ce-cosmos/Turkish-Llama-8b-Instruct-v0.1}
\cite{kesgin2024optimizing} and the results are compared in Table \ref{tab:zero_shot_llm}. Neither model is fine-tuned on
StutterStigma-TR, and no examples from the dataset are included in
the prompts. The prompts are fixed before evaluation and are not
selected using the test labels. Generation is deterministic using
greedy decoding. Model outputs are parsed automatically. Outputs that cannot be mapped
to one of the requested class labels are retained as invalid responses
and are not manually corrected. Response coverage is 0.9900 for Qwen2.5-7B and 0.9925 for
Turkish-Llama-8B in the three-class task, and 0.9942 and 0.9967,
respectively, for the dedicated binary prompts. The reported Macro-F1 values are calculated over all 1,204 test samples from StutterStigma-TR dataset.
Invalid model outputs are treated as incorrect predictions rather than being
excluded from the evaluation.

The three-class prompt asks the model to
choose among Stigmatizing, Non-stigmatizing/contextual, and Supportive.
A separate binary prompt asks for either Stigmatizing or Non-stigma. The prompts contain the corresponding class definitions and decision
rules, including guidance for reported or quoted harmful language,
neutral references, personal experience, and supportive or
anti-stigma content, but contain no labelled examples.
The human annotations remain the reference labels for both tasks.

\begin{table}[!t]
\caption{Comparison of the Proposed StutterStigma-BERT Model with Two Zero-Shot LLM Models Evaluated on StutterStigma-TR Held-out Test Set}
\label{tab:zero_shot_llm}
\centering
\scriptsize
\setlength{\tabcolsep}{2.2pt}
\renewcommand{\arraystretch}{1.05}

\begin{adjustbox}{max width=\columnwidth}
\begin{tabular}{lccc}
\toprule
\textbf{Task} &
\makecell{\textbf{Qwen2.5-7B}\\\cite{qwen2024qwen25}} &
\makecell{\textbf{Turkish-Llama-8B}\\\cite{kesgin2024optimizing}} &
\makecell{\textbf{StutterStigma-BERT (Ours)}\\
\textbf{(20-seed mean)}} \\
\midrule

Three-class Macro-F1 &
43.20\% &
24.61\% &
\textbf{73.54\%} \\

Binary Macro-F1 &
67.36\% &
71.74\% &
\textbf{85.17\%} \\

\bottomrule
\end{tabular}
\end{adjustbox}

\vspace{1mm}
\begin{minipage}{0.95\columnwidth}
\footnotesize
\textit{Note.} Values are reported as Macro-F1 percentages. LLM results
are obtained without fine-tuning, few-shot examples, or prompt selection
using gold test labels. Invalid outputs are not manually corrected. The
binary values come from the dedicated binary prompt. The supervised
reference values correspond to the task-matched configurations of the proposed
StutterStigma-BERT model, summarized
across 20 random seeds, and are included only as descriptive reference
points.

\end{minipage}
\end{table}

Both LLMs obtain higher Macro-F1 for the binary classification task (67.36\% for Qwen2.5-7B and 71.74\% for Turkish-Llama-8B), but the strongest zero-shot binary result still remains below the performance of our proposed supervised binary model. The gap between the zero-shot LLM models and the proposed StutterStigma-BERT is especially large for the three-class task, where the model must separate contextual and supportive uses in addition to identifying
stigma. These results are limited to the two models and prompts evaluated here and do not show that zero-shot LLMs in general are unsuitable for this task. Under this fixed zero-shot setup, the selected instruction-tuned LLMs achieve
lower Macro-F1 than the corresponding supervised StutterStigma-BERT
systems trained on the annotated StutterStigma-TR dataset.

\FloatBarrier
\subsection{Error Analysis on the Proposed StutterStigma-BERT}
\label{sec:error_analysis}

Error recurrence is examined across the same 20 random seeds used in the
final robustness analyses. The analysis therefore uses predictions from the proposed StutterStigma-BERT for both three-class and binary classification tasks, and the results are presented in Table \ref{tab:recurrent_errors}. For the
binary task, each seed uses its corresponding validation-selected decision
threshold. A test post is defined as a \textit{hard recurrent error} when it is
misclassified in at least 15 of the 20 seeds, while posts misclassified in all
20 seeds are considered persistent errors.

\FloatBarrier

\begin{table}[!ht]
\caption{
Recurrent Test-Set Errors Across the Final 20-Seed Analyses
}
\label{tab:recurrent_errors}
\centering
\small
\setlength{\tabcolsep}{5pt}

{
\begin{tabular}{lcc}
\toprule
\textbf{Error recurrence} &
\textbf{Three-class} &
\textbf{Binary} \\
\midrule

At least 1 of 20 seeds  & 533 & 333 \\
At least 5 of 20 seeds  & 386 & 215 \\
At least 10 of 20 seeds & 308 & 161 \\
At least 15 of 20 seeds & 244 & 137 \\
All 20 seeds            & 123 & 69 \\

\bottomrule
\end{tabular}

\vspace{1mm}
\begin{minipage}{0.95\columnwidth}
\footnotesize
\textit{Note.} Both tasks use the proposed StutterStigma-BERT
model. Binary predictions use the
validation-selected threshold associated with each seed. Counts represent
unique posts meeting the stated error-recurrence criterion.
\end{minipage}
}
\end{table}

For the three-class task, 533 of the 1,204 held-out posts are
misclassified in at least one seed, whereas 244 meet the hard recurrent-error
criterion and 123 are misclassified in all 20 seeds. To summarize the
direction of the hard recurrent errors, each of the 244 posts is assigned to
its most frequent incorrect predicted class across the 20 seeds. The most
common direction is
\textit{Non-stigmatizing/contextual} $\rightarrow$ \textit{Supportive}
($n=60$), followed by
\textit{Stigmatizing} $\rightarrow$ \textit{Non-stigmatizing/contextual}
($n=56$) and
\textit{Supportive} $\rightarrow$ \textit{Non-stigmatizing/contextual}
($n=45$). The remaining recurrent directions are
\textit{Non-stigmatizing/contextual} $\rightarrow$ \textit{Stigmatizing}
($n=37$),
\textit{Stigmatizing} $\rightarrow$ \textit{Supportive}
($n=30$), and
\textit{Supportive} $\rightarrow$ \textit{Stigmatizing}
($n=16$).
These error directions show that recurrent three-class errors occur on both
sides of the stigma decision boundary and also within the two non-stigmatizing
discourse categories. In particular, 105 of the 244 hard recurrent errors
involve confusion between \textit{Non-stigmatizing/contextual} and
\textit{Supportive} posts, illustrating the additional stance distinction
required by the three-class formulation.

For the binary task, 333 test posts are misclassified in at least one seed,
137 are hard recurrent errors, and 69 are misclassified in all 20 seeds.
Among the 137 hard recurrent binary errors, 87 are false negatives
(\textit{Stigmatizing} predicted as \textit{Non-stigma}) and 50 are false
positives. Among the 69 persistent errors, 45 are false negatives and
24 are false positives.
Because the binary \textit{Non-stigma} class combines the original
\textit{Non-stigmatizing/contextual} and \textit{Supportive} classes, the
50 hard recurrent false positives can also be examined according to their
original three-class labels. Thirty-eight originate from
\textit{Non-stigmatizing/contextual} posts and 12 from \textit{Supportive}
posts. Among the 24 false positives misclassified in all 20 seeds,
21 are originally \textit{Non-stigmatizing/contextual} and three are
\textit{Supportive}.

The two tasks share a substantial subset of recurrently difficult examples.
A total of 115 posts are hard recurrent errors in both tasks. An additional
129 are hard recurrent errors only in the three-class task, whereas 22 are
hard recurrent errors only in the binary task. The remaining 938 test posts
do not meet the hard recurrent-error criterion in either task. The larger
three-class-specific set is consistent with the additional requirement to
distinguish supportive stance from non-stigmatizing or contextual mention,
rather than only separating stigma from non-stigma.

\section{Discussion }

This study introduces a condition-specific framework for detecting
stuttering-related stigma in Turkish social-media posts. 
The results support one central point: in StutterStigma-TR the label depends on
how a post positions the stuttering-related expression, not on whether the
expression occurs. Every post contains such an expression, yet the same words
appear in mockery, clinical description, self-disclosure, reported insults, and
counter-speech. Models that read only surface negativity therefore transfer
poorly, while models trained on the condition-specific labels perform
substantially better. In the following, we will discuss more on the experiment results and observations from the presented work.

\subsection{From Target Presence to Target Function}

The matched Raw BERTurk comparison provides a direct view of how explicit
target information contributes to classification. In the three-class task,
mean internal-validation Macro-F1 increases from 75.01\% for Raw BERTurk
to 75.37\% with explicit target marking. This pattern indicates that explicitly
identifying the relevant stuttering-related expression contributes additional
information beyond the unmarked contextual representation. In the binary task,
the corresponding validation Macro-F1 increases from 84.88\% to 85.59\%. However, the differences among
the stronger target-aware binary configurations remain relatively small.
The contribution of explicit target information is therefore not identical
across the two classification settings and should be interpreted together
with the subsequent target-aware configurations rather than from target
marking alone.
The stronger target-aware configurations are relatively close to one another
in both tasks. This pattern is consistent with target-dependent and
aspect-based modelling, in which interpretation depends on the relation
between a target and its surrounding context
\cite{tang_etal_2016,ma2017interactive,sun_etal_2019,wu_ong_2021}.
In the present setting, however, the results do not support a monotonic gain
from each additional target-aware component.

This distinction is consistent with previous work on stuttering-related social media content. Donaher and Minkoff \cite{donaher2014portrayal} show that YouTube content about stuttering includes educational material, entertainment, self-disclosure, misinformation, and negative comments. Çavdar et al. \cite{ccavdar2025reliable} similarly show that Turkish YouTube and Instagram content varies in reliability, quality, source type, and engagement. On text-based platforms, Azios et al. \cite{azios2022twitter} and Rasoli Jokar et al. \cite{rasoli2025stuttering} report that stuttering may appear in negative, neutral, supportive, political, and awareness-oriented discourse. These studies show why stuttering-related posts cannot be treated as a single discourse type. The disability-related NLP literature provides a parallel concern. Prior studies show that disability-related terms may be associated with higher negativity or toxicity scores even when they appear in neutral, self-descriptive, or advocacy-oriented language \cite{hutchinson2020social, hassan_etal_2021, venkit_etal_2023}. At the same time, subtle ableist meanings may be missed when they do not contain direct insult, profanity, threat, or other conventional markers of harmful language \cite{phutane2025cold, rizvi2025autalic, rizvi2025beyond}. This creates a difficult boundary: systems need to avoid treating disability-related language as harmful by default, but they also need to detect harm when it is indirect or socially normalized.

The repeated-run error analysis shows this boundary clearly. Some
supportive or contextual posts remain difficult because they contain
quoted stigma, negative language, or references to discrimination.
Conversely, some stigmatizing posts are repeatedly missed when the
harm is indirect, especially when stuttering is linked to competence,
credibility, or suitability for speaking roles. These errors are not
mainly about finding the target expression. They arise from how the
target is framed in the post. Condition-specific work on anti-autistic ableist language is especially relevant here. AUTALIC and related studies are developed because autism-related language often requires context-preserving annotation and because harm may depend on how the target group is framed rather than on the presence of an identity-related term alone \cite{rizvi2025autalic, rizvi2025beyond}. The present study applies the same broader logic to stuttering-related discourse in Turkish. The specific stereotypes differ, but the modelling challenge is similar: the system must distinguish harmful framing from neutral mention, self-reference, reported speech, and supportive response. The direct-transfer results in the present study point in the same
direction. Models trained for broader hate speech, bullying, and
toxicity detect some harmful content, but they remain well below the
condition-specific supervised systems.

This is why the three-class formulation is useful. A binary stigma versus non-stigma task can support applied detection, but it collapses different non-stigmatizing uses of stuttering-related language. A clinical mention, a personal disclosure, a quoted example of stigma, and an anti-stigma response may all be non-stigmatizing, but they are not the same kind of post. Keeping Stigmatizing, Non-stigmatizing/contextual, and Supportive separate allows the evaluation to test whether models can distinguish harmful use from neutral or supportive discourse.

\subsection{Modeling the Target Context Relation}

The model comparisons show that the relevance of target-specific information
depends on the granularity of the classification problem. Raw BERTurk
substantially outperforms the lexical TF--IDF baselines in both tasks,
confirming the importance of contextual transformer representations
\cite{devlin2019bert,schweter_2020_berturk}. The common-protocol comparison
then isolates the additional information provided by explicit target-aware
representations. Designing different components in our proposed StutterStigma-BERT model, it outperforms raw BERTurk model by increasing the mean internal-validation Macro-F1 from 75.01\% to 76.32\% for three-class classification and from 84.88\% to 86.18\% for binary classification.
This pattern is consistent with target-dependent and aspect-based modeling,
where interpretation depends on the relation between a particular target and
its surrounding context rather than on the sentence alone
\cite{tang_etal_2016, ma2017interactive, sun_etal_2019, wu_ong_2021}.
The present task differs from conventional aspect-based sentiment analysis
because the target is not a product feature or opinion aspect. It is a
stuttering-related expression whose social meaning depends on how it is used.
Nevertheless, the same general modeling principle applies: the classifier
benefits from knowing which span is relevant and from representing that span
together with its surrounding context.

In our proposed model, explicit target marking provides a structural cue identifying the
stuttering-related expression within the input sequence. Target-span pooling
then creates a dedicated contextual representation of the marked span, while
target-conditioned attention represents surrounding non-target tokens in
relation to that target. These mechanisms are particularly relevant because
the same stuttering-related expression may occur in mockery, clinical
description, personal experience, reported speech, political insult, or
supportive discourse.
The OrthoTarget configurations additionally incorporate character-level and
handcrafted orthographic information. Such information is relevant to social
media because stigma may be expressed through repeated letters, stylized
imitation, punctuation, capitalization, laughter markers, profanity, or
non-standard spelling. Character-level modelling can capture spelling-level
and subword variation in noisy text
\cite{zhang_etal_2015_charcnn}. These surface cues complement rather than
replace contextual semantic representations, because many stigmatizing posts
depend on indirect evaluation or discourse framing rather than conspicuous
orthographic patterns.
Whole components together in StutterStigma-BERT model
extend the semantic representation with additional global--attended-context interactions and adds gated global and target-local discourse features. 


\subsection{Why Discourse Remains Difficult}

The remaining errors indicate that the most difficult cases require more
than recognizing a stuttering-related expression or detecting overtly
negative language. Classification depends on discourse function and stance:
whether a post is endorsing, reporting, quoting, rejecting, normalizing, or
otherwise contextualizing potentially stigmatizing content. This distinction
is especially important for stuttering because stigma can extend beyond
direct ridicule to assumptions about competence, credibility, confidence,
intelligence, emotional control, and suitability for communication-intensive
roles \cite{boyle2013assessment, ozdemir2011public, st_louis_2015,
tichenor2019stuttering}.

The final 20-seed error analysis illustrates this difficulty at the class
boundaries. In the three-class task, 244 posts are misclassified in at least
15 of the 20 seeds. Of these, 105 involve confusion specifically between
\textit{Non-stigmatizing/contextual} and \textit{Supportive} discourse:
60 are predominantly predicted as \textit{Supportive} despite a
\textit{Non-stigmatizing/contextual} gold label, and 45 show the reverse
pattern. This indicates that distinguishing active supportive stance from
neutral, descriptive, or contextual mention remains an important source of
recurrent error.
The binary analysis shows a related boundary problem. Among the 137 hard
recurrent binary errors, 50 are false positives. Thirty-eight of these
originate from the original \textit{Non-stigmatizing/contextual} class and
12 from the \textit{Supportive} class. Among the 24 false positives that
persist across all 20 seeds, 21 are originally
\textit{Non-stigmatizing/contextual} and three are \textit{Supportive}.
Thus, recurrent over-classification is concentrated primarily in
non-stigmatizing contextual discourse, although supportive posts also
contribute to this error type.

Reported or quoted stigma provides one plausible source of this ambiguity.
A post may reproduce harmful wording in order to describe an event, report
another person's statement, criticize mockery, or oppose a stereotype. In
such cases, the presence of stigma-related wording does not by itself
determine the discourse function. The distinction instead depends on the
speaker's position toward the referenced content. Supportive counter-speech
poses a similar challenge because harmful language may be repeated in order
to reject it or defend people who stutter. This broader problem is consistent
with disability-related NLP studies showing that disability-related language
can be over-associated with negativity or toxicity when context and stance
are insufficiently represented
\cite{hutchinson2020social, hassan_etal_2021, venkit_etal_2023}.

The opposite error also remains substantial. Of the 137 hard recurrent
binary errors, 87 are false negatives, and 45 of the 69 errors that persist
across all 20 seeds are Stigmatizing posts repeatedly predicted as
Non-stigma. These cases show that harmful meaning is not always accompanied
by conspicuous lexical or orthographic cues. Stuttering-related stigma may
be conveyed indirectly through implication, comparison, lowered
expectations, or assumptions about communicative competence and social
suitability rather than through an explicit insult.
This interpretation is consistent with broader work on ableist harm.
Disability-related harm can be indirect, humorous, paternalistic, or socially
normalized, and may occur through stereotypes, lowered expectations,
exclusionary assumptions, or apparently ordinary comparisons rather than
through conventional hate-speech markers
\cite{hutchinson2020social, venkit_etal_2023, phutane2025cold,
rizvi2025autalic, rizvi2025beyond}. For stuttering-related discourse, the
relevant distinction is therefore not simply whether a negative expression
is present, but how disfluent speech and people who stutter are positioned
within the surrounding discourse.

Some difficulty also reflects the limits of post-level classification.
Social-media posts can be short, ironic, elliptical, or dependent on earlier
interaction. The models evaluate the available post text but do not
systematically incorporate the full conversational thread, speaker history,
or wider interactional context. A post that appears harmful in isolation may
be reporting or criticizing stigma, whereas apparently neutral wording may
depend on prior context to convey a stigmatizing implication.
The recurrent-error overlap across tasks further shows that some examples
remain difficult even after the label space is simplified. A total of 115
test posts are hard recurrent errors in both the three-class and binary
tasks. Another 129 are hard recurrent only in the three-class task, whereas
22 are hard recurrent only in the binary task. The remaining 938 posts do
not meet the hard recurrent-error criterion in either task. The larger
three-class-specific error set is consistent with the additional requirement
to distinguish \textit{Supportive} discourse from
\textit{Non-stigmatizing/contextual} discourse.

Overall, the recurrent errors point to stance, quotation, implication, and
wider discourse context as continuing challenges. These findings also
support the use of the system as a research and human-review aid rather than
as a substitute for contextual human interpretation in ambiguous cases.

\subsection{Implications for Disability-Informed NLP and Stuttering Research}

The findings have implications for both disability-informed NLP and stuttering research. For NLP, the study shows the value of developing condition-specific models for ableist harm. Ableism is not expressed in the same way across all disability groups. Different conditions are associated with different stereotypes, linguistic forms, and social expectations. Work on anti-autistic ableist language already shows the need for condition-specific annotation and evaluation when harmful meaning depends on how a group is framed in context \cite{rizvi2025autalic, rizvi2025beyond}. The present study extends this direction to stuttering-related discourse in Turkish.

For stuttering, the relevant social meanings are closely tied to speech fluency and communicative participation. Stigma may appear through direct mockery or written imitation, but it may also appear through assumptions about competence, credibility, confidence, or suitability for speaking roles. A model designed for this task therefore needs to recognize not only harmful language in a general sense, but also the particular ways in which stuttering is positioned socially. The target-aware framework addresses this need by keeping the stuttering-related expression central to the classification problem while allowing the label to depend on context and stance. The study also highlights the importance of annotation design. The annotation process combines speech-language therapy expertise with lived-experience input from people who stutter. This is important because the task involves subtle distinctions among stigma, neutral mention, reported discrimination, quotation, and support. In disability-related NLP, labels should not be treated as purely technical outputs detached from the communities being represented. Lived-experience involvement does not eliminate ambiguity, but it can make the annotation process more sensitive to meanings that may be missed by general sentiment or toxicity frameworks.

For stuttering research, the proposed framework offers a way to examine public representations of stuttering at scale. Previous studies show that online stuttering-related content includes education, misinformation, awareness, support, negative stereotyping, and variable source quality \cite{donaher2014portrayal, azios2022twitter, ccavdar2025reliable, rasoli2025stuttering}. A model-based approach can complement this work by helping researchers identify broader patterns in how stigmatizing, contextual, and supportive posts are distributed across large social media datasets. This content-focused approach can also be connected with research on the social media experiences of people who stutter. Çavdar et al. \cite{ccavdar2026stuttering} examine how people who stutter experience and navigate social media, including issues related to self-presentation, interaction, and cyberbullying concerns. The present study addresses the public discourse side of the same environment: the kinds of stuttering-related meanings that people who stutter may encounter online. Together, these lines of work can help connect online representation with participation, avoidance, advocacy, and perceived safety.

The proposed framework is best suited for research and auditing rather than fully automated intervention. It can support large-scale monitoring of stuttering-related discourse, help identify examples for qualitative analysis, and assist human review in high-volume settings. It may also be useful for tracking how stigma, support, and awareness change over time or across platform contexts. These uses preserve space for human interpretation, which remains necessary in cases involving quotation, counter-speech, self-disclosure, or ambiguous stance.

\subsection{Limitations and Future Work}
This study has several limitations. First, \textit{data scope}: StutterStigma-TR
was built by targeted retrieval of public Turkish X/Twitter posts containing
predefined stuttering-related expressions. It is a purposive, condition-specific
sample from one platform and period, so its class distribution does not estimate
the prevalence of stuttering-related stigma, and stigma expressed without a
detectable target expression (e.g., euphemisms, memes, screenshots, or audio and
video imitation) falls outside the present approach.

Second, \textit{language and context}: the study is limited to Turkish, and its
annotation scheme, target patterns, and results may not transfer directly to
other languages, in which stigma may rely on different metaphors, humour, and
stereotypes. The models also classify individual posts without access to the
conversation thread, reply structure, or speaker history, which limits their
handling of quotation, reported stigma, sarcasm, and counter-speech.

Third, \textit{annotation}: outside the 645-post reliability subset, each post
was initially labelled by a single annotator, with consensus review limited to
flagged or disputed cases, so some label noise cannot be excluded. Judgements of
stigma also vary across individuals; although the team included a person who
stutters and an independent lived-experience review was conducted, larger and
more diverse panels would allow disagreement to be studied as meaningful
variation. The stigma subtypes were not tested for reliability and are used only
descriptively.

Fourth, \textit{evaluation}: all results use one fixed held-out split, and the
20-seed analysis measures sensitivity to random initialization rather than
external validity. The backbone comparison is restricted to three seeds and
requires backbone-specific adaptations, so it provides descriptive evidence
rather than a general ranking of model families. The validation-selected binary
threshold varies substantially across seeds and should not be read as a
calibrated or universally optimal operating point. The zero-shot evaluation
covers only two models and fixed prompts, and the error patterns were
identified without independent coding.

Future work should therefore evaluate the framework on independently collected,
temporal, cross-platform, and multimodal data; incorporate conversational
context; model stigma subtypes with reliability-tested annotations; involve
broader lived-experience panels; and extend the comparison to additional Turkish
and multilingual backbones and prompting strategies.



\section{Conclusion}

This study introduced StutterStigma-TR, a condition-specific Turkish dataset of 6,019 social media posts, together with StutterStigma-BERT, a target-aware framework for distinguishing stigmatizing, non-stigmatizing/contextual, and supportive stuttering-related discourse. The framework integrates an explicit representation of the stuttering-related target with contextual, character-level, orthographic, and discourse-level information. The proposed architecture architecture trained with cross-entropy loss as the final configuration for both the three-class and binary tasks. Across 20 random seeds, the final models achieved mean Macro-F1 scores of 73.54\% for the three-class task and 85.17\% for the binary task. For the primary three-class task, we evaluated different backbone models and the results show that the StutterStigma-BERT model absed on BERTurk backbone achieved higher internal-validation Macro-F1 than the corresponding TurkishBERTweet- and Turkish Mamba-based implementations under matched configurations. Overall, the findings indicate that detecting stuttering-related stigma requires interpreting how a stuttering-related expression is used in context, rather than merely recognizing the presence of the expression itself. StutterStigma-TR and StutterStigma-BERT provide a condition-specific resource and modelling framework to support research on disability-related online discourse, discourse auditing, and human-reviewed content analysis, rather than fully automated moderation. Future work should evaluate the framework on independent, cross-platform, multilingual, and multimodal data and should incorporate broader conversational context and more extensive lived-experience evaluation.


\section*{Data Availability and Privacy Statement}
The study used publicly accessible Turkish-language posts collected from
X/Twitter. As part of dataset preparation, usernames and URLs
were replaced with the standardized tokens \texttt{[USER]} and
\texttt{[URL]}, respectively, before model development. User-identifying
account information was not used as a model feature. The article reports aggregate statistics, model results, and
error patterns rather than reproducing identifiable user posts. All example
posts shown in the figures are constructed illustrative examples and do not
reproduce identifiable user content. To request access to the StutterStigma-TR dataset, please contact the authors of the paper.


\bibliographystyle{unsrt}  
\bibliography{references}

\end{document}